\documentclass[10pt,twocolumn,letterpaper]{article}

\usepackage{cvpr}
\usepackage{times}
\usepackage{epsfig}
\usepackage{graphicx}
\usepackage{amsmath}
\usepackage{amssymb}
\usepackage{mathtools}
\usepackage{tcolorbox}
\usepackage{tabu}
\usepackage{empheq}

\usepackage{placeins}
\usepackage{afterpage}

\usepackage[format=plain,labelformat=simple,labelsep=period,font=small,skip=4pt,compatibility=false]{caption}
\usepackage[font=scriptsize,skip=2pt]{subcaption}

\usepackage{tabularx}
\usepackage{setspace}
\usepackage{nicefrac}
\usepackage{booktabs}
\usepackage[super]{nth}

\usepackage{siunitx}
\usepackage[pagebackref=true,breaklinks=true,letterpaper=true,colorlinks,bookmarks=false]{hyperref}

 \cvprfinalcopy 

\def\cvprPaperID{1969} 

\ifcvprfinal\fi

\newcommand{\qed}{\nobreak \ifvmode \relax \else
      \ifdim\lastskip<1.5em \hskip-\lastskip
      \hskip1.5em plus0em minus0.5em \fi \nobreak
      \vrule height0.75em width0.5em depth0.25em\fi}

\title{Video Extraction from a Single Blurry Image}
\title{Learning to Extract a Video Sequence from a Single Motion-Blurred Image}

\author{Meiguang Jin\quad Givi Meishvili\quad Paolo Favaro\\
University of Bern, Switzerland\\
{\tt\small \{jin, meishvili, favaro\}@inf.unibe.ch}
}

\begin{document}

\twocolumn[{%
\renewcommand\twocolumn[1][]{#1}%
\maketitle
	\vspace{-.5cm}
   	\begin{minipage}{0.124\textwidth}
	\includegraphics[width=\linewidth,trim={5cm 2cm 2cm 3cm},clip]{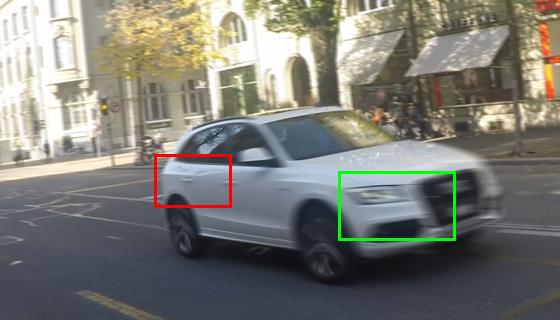}%
	\end{minipage}~%
   	\begin{minipage}{0.124\textwidth}
	\includegraphics[width=\linewidth,trim={5cm 2cm 2cm 3cm},clip]{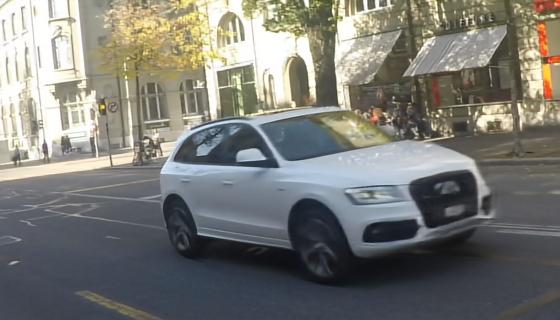}%
	\includegraphics[width=\linewidth,trim={5cm 2cm 2cm 3cm},clip]{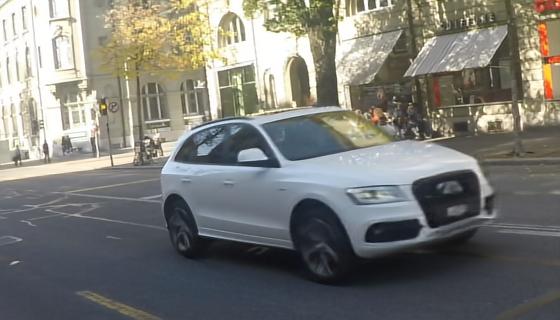}%
	\includegraphics[width=\linewidth,trim={5cm 2cm 2cm 3cm},clip]{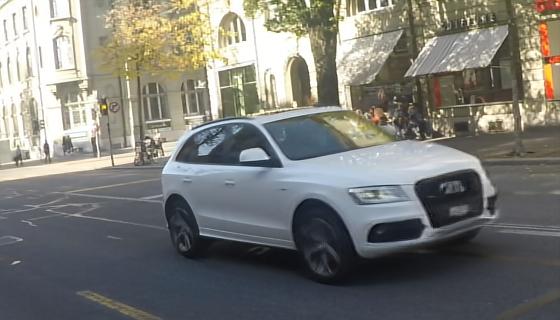}%
	\includegraphics[width=\linewidth,trim={5cm 2cm 2cm 3cm},clip]{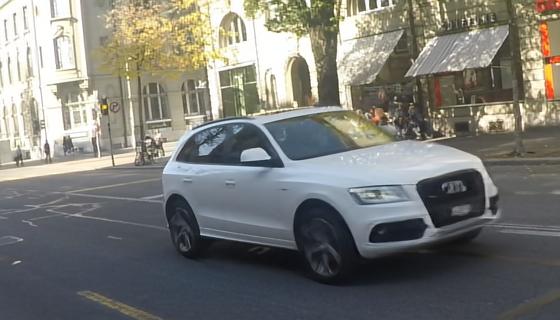}%
	\includegraphics[width=\linewidth,trim={5cm 2cm 2cm 3cm},clip]{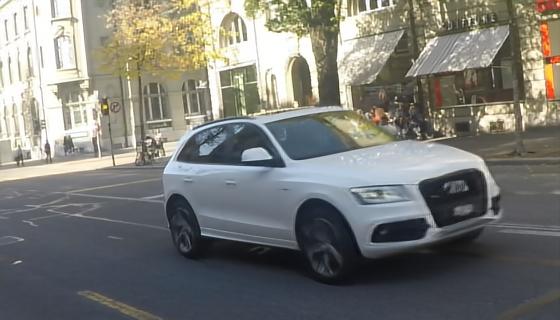}%
	\includegraphics[width=\linewidth,trim={5cm 2cm 2cm 3cm},clip]{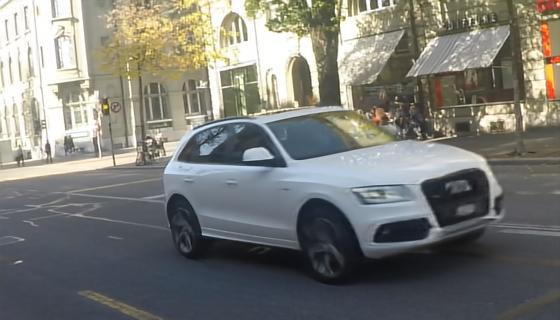}%
	\includegraphics[width=\linewidth,trim={5cm 2cm 2cm 3cm},clip]{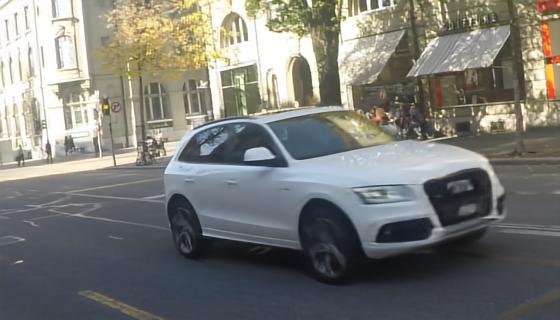}%
	\end{minipage}\\
   	\begin{minipage}{0.121\textwidth}
	\setlength{\fboxsep}{0pt}
	\setlength{\fboxrule}{1pt}
	\fcolorbox{red}{white}{\includegraphics[width=\linewidth,angle=-0,origin=c]{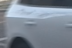}}%
	\end{minipage}\hspace{1.4mm}%
   	\begin{minipage}{0.124\textwidth}
	\includegraphics[width=\linewidth,angle=-0,origin=c]{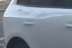}%
	\includegraphics[width=\linewidth,angle=-0,origin=c]{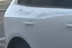}%
	\includegraphics[width=\linewidth,angle=-0,origin=c]{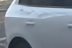}%
	\includegraphics[width=\linewidth,angle=-0,origin=c]{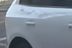}%
	\includegraphics[width=\linewidth,angle=-0,origin=c]{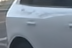}%
	\includegraphics[width=\linewidth,angle=-0,origin=c]{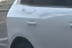}%
	\includegraphics[width=\linewidth,angle=-0,origin=c]{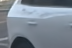}
	\end{minipage}\\
   	\begin{minipage}{0.121\textwidth}
	\setlength{\fboxsep}{0pt}
	\setlength{\fboxrule}{1pt}
	\fcolorbox{green}{white}{\includegraphics[width=\linewidth,angle=-0,origin=c]{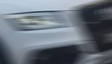}}%
	\end{minipage}\hspace{1.4mm}%
   	\begin{minipage}{0.124\textwidth}
	\includegraphics[width=\linewidth,angle=-0,origin=c]{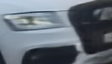}%
	\includegraphics[width=\linewidth,angle=-0,origin=c]{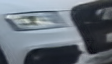}%
	\includegraphics[width=\linewidth,angle=-0,origin=c]{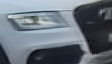}%
	\includegraphics[width=\linewidth,angle=-0,origin=c]{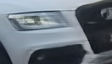}%
	\includegraphics[width=\linewidth,angle=-0,origin=c]{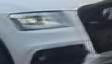}%
	\includegraphics[width=\linewidth,angle=-0,origin=c]{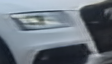}%
	\includegraphics[width=\linewidth,angle=-0,origin=c]{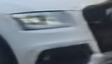}
	\end{minipage}
    \captionof{figure}{Multiple frames extracted from a single motion blurred image. On the left column we show the input image and two enlarged details with different motion blur. On the columns to the right we show the estimated $7$ frames and corresponding enlargements. 
  \label{fig:teaser}}\vspace{.5cm}
}]

\maketitle

\begin{abstract} 
We present a method to extract a video sequence from a single motion-blurred image. 
Motion-blurred images are the result of an averaging process, where instant frames are accumulated over time during the exposure of the sensor. 
Unfortunately, reversing this process is nontrivial.
Firstly, averaging destroys the temporal ordering of the frames. 
Secondly, the recovery of a single frame is a blind deconvolution task, which is highly ill-posed. 
We present a deep learning scheme that gradually reconstructs a temporal ordering by sequentially extracting pairs of frames.
Our main contribution is to introduce loss functions invariant to the temporal order.
This lets a neural network choose during training what frame to output among the possible combinations.
We also address the ill-posedness of deblurring by designing a network with a large receptive field and implemented via resampling to achieve a higher computational efficiency.
Our proposed method can successfully retrieve sharp image sequences from a single motion blurred image and can generalize well on synthetic and real datasets captured with different cameras.
\end{abstract}

\section{Introduction} 



It is often said that photos capture a memory, an instant in time. Technically, however, this is not strictly true. Photos require a finite exposure to accumulate light from the scene. Thus, objects moving during the exposure generate motion blur in a photo. 
Motion blur is an image degradation that makes visual content less interpretable and is therefore often seen as a nuisance. However, motion blur also combines information about both texture and motion of the objects in a single blurry image. Hence, recovering texture and motion from motion blurred images can be used to understand the dynamics of a scene (\eg, in entertainment with sports or in surveillance when monitoring the traffic). 
The task of recovering a blur kernel and a sharp image, whose convolution gives rise to a given blurry image, is called \emph{motion deblurring} or \emph{blind deconvolution}.
Unfortunately, this formulation of the task is accurate only for some special cases of motion blur. In particular, it holds in the cases when blur is the same across an image (the so-called shift invariant blur \cite{Lai_2016_CVPR}) or when blur can be modeled as a linear combination of a basis of shift fields (\eg, in the case of camera shake \cite{Fergus_2006_SIGGRAPH}).
However, in the case of multiple moving objects, also called \emph{dynamic blur} \cite{Nah_2017_CVPR}, a blurry image is no longer some convolution of a blur pattern with a single sharp image. In this case, a blurry image is the averaging over time of instant frames, where multiple objects move independently and cause occlusions. 

In this paper we introduce blind deconvolution with dynamic blur as the task of recovering a sequence of sharp frames from a single blurry image. As illustrated in Fig.~\ref{fig:teaser}, given a single motion-blurred image (left column) we aim at recovering a sequence of $7$ frames each depicting some instantaneous motion of the objects in the scene. To the best of our knowledge, this is the first time this problem has been posed and addressed. 
The two main challenges in solving this task are that: 1) blur removal is an ill-posed problem and 2) averaging over time destroys the temporal ordering of the instant frames. To handle the ill-posedness of deblurring we use a deep learning approach and train a convolutional neural network with a large receptive field. A large receptive field could be achieved by using large convolutional filters. However, such filters would have a detrimental impact on the memory requirements and the computational cost of the network. We avoid these issues by using a re-sampling layer (see Sec.~\ref{sec:implementation}).
Handling the loss of the temporal ordering is instead a less well-studied problem in the literature. To make matters worse, this ordering ambiguity extends to the motion of each object in the scene, thus leading to a combinatorial explosion of valid solutions. One possible exception to this scenario is the estimation of the frame in the  middle of the sequence. In most motion blurred images the middle frame corresponds to the center of mass of the local blur, which can be unambiguously identified given the blurry input image \cite{Nah_2017_CVPR,debluringWild_2017_GCPR}. However, as we show in the Experiments section, the other frames do not enjoy the same uniqueness. We find that training a neural network by defining a loss on a specific frame of the sequence, other than the middle one, yields very poor results (see Sec.~\ref{sec:experiments}).
We thus analyze temporal ambiguities in Sec.~\ref{sec:ambiguities} and present a novel deep learning method that extracts instant frames in a sequential manner. 
Our main contribution is to train neural networks via loss functions that are invariant to the temporal ordering of the frames. These loss functions use the average of two frames and the absolute value of their difference as targets. This allows each network to choose which frames to output during training. Moreover, to make the network outputs more realistic and sharp, we use adversarial training \cite{Goodfellow_2014_NIPS}. 
In the Experiments section we demonstrate that our trained networks can successfully extract videos from both synthetic and real motion-blurred images. In addition to providing accurate motion information about objects in the scene, we plan to use our method for video editing and temporal superresolution of videos. By exploiting the information embedded in motion blur, our method has the potential to interpolate subsequent frames with high accuracy.

\section{Prior Work} 
\noindent\textbf{Uniform Motion Deblurring.} Blind deconvolution has been studied for several decades, and tremendous progress has been made in the case of uniform motion blur  \cite{Pan_2016_CVPR, Michaeli_2014_ECCV,Chakrabarti_2016_ECCV,Dong_2017_ICCV,Yan_2017_CVPR} and camera shake blur \cite{Hirsh, Wipf_2013_NIPS, Bahat_2017_ICCV}. Yan ~\etal \cite{Yan_2017_CVPR} present an extremely effective image prior by combining the bright and dark channel priors of Pan \etal~\cite{Pan_2016_CVPR}. 
Michaeli \etal~\cite{Michaeli_2014_ECCV} incorporated recurrence of small image patches across different scales of a natural image. Gong ~\etal \cite{Gong_2016_CVPR} introduced a gradient activation algorithm 
for kernel estimation. 

\noindent\textbf{Non-Uniform Motion Deblurring.} Recently, the general motion deblurring problem has attracted a lot of attention. 
\cite{Kim_2013_ICCV} propose an energy model to estimate different motion blurs and their associated pixel-wise weights. \cite{Kim_2014_CVPR} use a TV-L1 model to simultaneously estimate  motion flow and a latent sharp image. 
Sun ~\etal \cite{Sun_2015_CVPR} trained a convolutional neural network (CNN) for predicting a probability distribution of motion blurs. 
A sharp image is estimated 
by using a patch-level image prior. 
Pan \etal~\cite{Pan_2016_CVPR_1} developed an efficient algorithm to jointly estimate object segmentation and camera motion, where each layer is deblurred under the guidance of a soft-segmentation. \cite{Gong_2017_CVPR} estimated a dense motion flow with a fully convolutional neural network and recovered the latent sharp image from the estimated motion flow. 
Bahat \etal~\cite{Bahat_2017_ICCV} 
recover the unknown blur field by analyzing the spectral content and deblur the image from the estimated blur field with a patch recurrence prior.
Pan \etal~\cite{Pan_2017_ICCV} proposed a method to learn data fitting functions from a large set of motion blurred images with the associated ground truth blur kernels. 
Nimisha \etal~\cite{Nimisha_2017_ICCV} used adversarial training to learn blur-invariant features which fed to a decoder to produce a deblurred image.
Recent work \cite{Nah_2017_CVPR,debluringWild_2017_GCPR,Kim_2017_ICCV} suggests to generate synthetic data for dynamic scene motion blur by averaging consecutive frames captured with a high frame rate camera. This dataset could then be used to train a neural network. For example, \cite{Nah_2017_CVPR} trained an end-to-end model with a  multi-scale convolutional neural network to restore the latent image directly. 

\noindent\textbf{Video Deblurring.} A number of methods consider the task of restoring a sharp sequence from a blurry video sequence. Zhang \etal~\cite{Zhang_2015_CVPR} propose a video deblurring method that 
jointly estimates the motion between consecutive frames as well as blur within each frame. Sellent \etal~\cite{Sellent_2016_ECCV}  instead exploit a stereo video sequence. Wieschollek \etal~\cite{Wieschollek_2017_ICCV} introduce a recurrent network architecture to deblur images by taking temporal information into account. Kim \etal~\cite{Kim_2017_ICCV} also exploit a (spatio-temporal) recurrent network, but aim for real-time performance. Kim \etal~\cite{Kim2016} propose a method for simultaneously removing general blurs and estimating optical flow from a video sequence. Ren \etal~\cite{Ren_2017_ICCV} exploit semantic segmentation of each blurry frame to understand the scene contents and use different motion models for image regions to guide the optical flow estimation. Su \etal~\cite{Su_2017_CVPR} propose a CNN that deblurs videos by incorporating information accumulated across frames. Pan \etal~\cite{Pan_2017_CVPR} propose a framework to jointly estimate the scene flow and deblur the image. Park \etal~\cite{Park_2017_ICCV} develop a method for the joint estimation of camera pose, depth, deblurring, and super-resolution from a blurred image sequence.

However, none of these approaches solves the task of extracting a video sequence from \textit{a single} motion-blurred image. In the following sections, we first illustrate the main challenges of our problem, then we introduce our novel loss functions and show how they address these challenges. The network design is introduced in Sec.~\ref{sec:implementation} and tested on synthetic and real datasets in the Experiments section.


\section{From Video to Image}



An image $y\in\mathbf{R}^{M\times N}$ captured with exposure $\tau$ can be written as
\begin{equation}
			\textstyle y = g\left(\frac{1}{\tau} \int_{0}^{\tau} \tilde x(t) dt\right) = g\left(\frac{1}{T} \sum_{i=0}^{T-1} x[i]\right),
		\label{eq:forward}
\end{equation}
where $g$ is the camera response function, which relates the irradiance at the image plane to the measured image intensity, and $\tilde x(t)$ is the instant image (irradiance) at time $t$. We discretize the time axis into $T$ segments, and define a sequence of frames $x[i]$, with $i=1,\dots,T$. Each frame $x[i]$ corresponds to the integral of $\tilde x(t)$ over a segment, \ie,
\begin{align}
\textstyle x[i] = \frac{T}{\tau}\int_{\frac{\tau}{T}i}^{\frac{\tau}{T}(i+1)} \tilde x(t) dt.
\end{align}
Object motion introduces a relative shift (in pixels) of regions between subsequent instant images $\tilde x(t)$. Given the maximum shift $\Delta$ that we are interested in handling, and by defining negligible blur as a shift of $1$ pixel, we can define the maximum number $T$ of time segments by setting $T = \Delta$. This choice only ensures that on average each frame $x[i]$ will have no motion blur. However, motions with acceleration may cause blur larger than $1$ pixel in some frames.

The motion-blur model~\eqref{eq:forward} thus far described is quite general, as regions can shift in an unconstrained way and subsequent instant images can introduce or remove texture (occlusions). Indeed, this model can handle the most general case of motion-blur, often called \emph{dynamic blur}.
This suggests a new formulation of motion deblurring with dynamic blur:
\begin{quote} 
Given a motion-blurred image $y$, recover the $T$ frames $x[1], \dots, x[T]$ satisfying model~\eqref{eq:forward}.
\end{quote}
As mentioned in the Introduction, the task of recovering a sharp image from a blurry one is already known to be highly ill-posed. In our formulation, however, the task is made even more challenging by the loss of frame ordering in the model~\eqref{eq:forward}. It may be possible to determine the local ordering of subsequent frames by exploiting temporal smoothness, however, there exist several ambiguities that we describe and discuss in the next section.
For example, given $y$ it is impossible to know if the ordering of the original sequence was $x[1], \dots, x[T]$ or $x[T], \dots, x[1]$, which corresponds to all objects moving forward or backward in time. 
Due to the complexity of our task, we adopt a data-driven approach. We build a dataset of blurry images with corresponding ground truth frames by exploiting high frame-rate videos as in recent methods \cite{Nah_2017_CVPR,debluringWild_2017_GCPR}, and devise a novel training method with convolutional neural networks (see Sec.~\ref{sec:training}).



\begin{figure}
\centering
	\includegraphics[width=\linewidth,trim={2.5cm 0cm 2.5cm 0cm},clip]{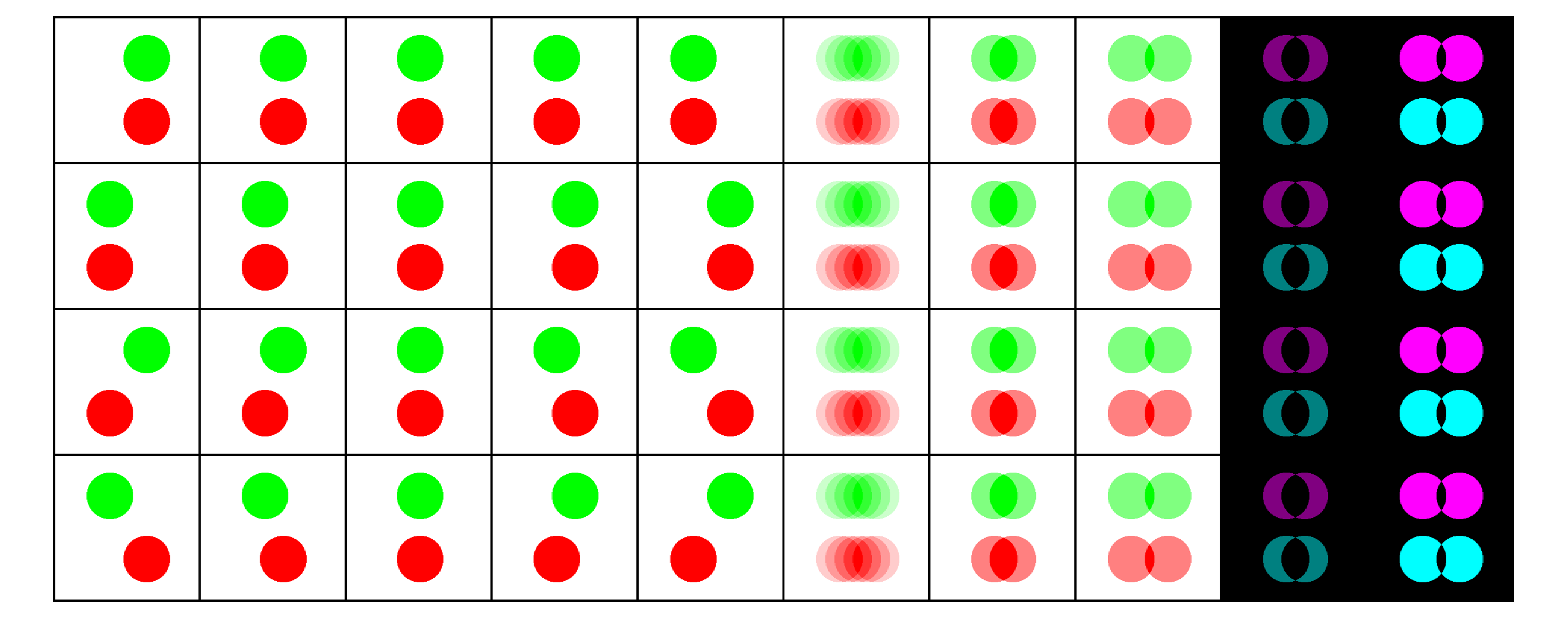}\\
	(a)\hspace{.45cm}(b)\hspace{.45cm}(c)\hspace{.45cm}(d)\hspace{.45cm}(e)\hspace{.45cm}(f)\hspace{.45cm}(g)\hspace{.45cm}(h)\hspace{.45cm}(i)\hspace{.45cm}(l)
	\caption{Temporal ordering ambiguities. In this toy example, we show two moving objects: a red and a green ball. Both are translating horizontally. Columns (a)-(e) show five video frames in four scenarios. Each of the four rows shows a plausible motion scenario of the two objects. Column (f) shows the blurry average of the first five columns. These averages are all identical, thus demonstrating that all four sequences are equally valid solutions. Column (g) shows the average of frame (b) and (d). Column
	(h) shows the average of frame (a) and (e). Column (i) shows the absolute difference of frame (b) and (d). Finally, column (h) shows the absolute difference of frame (a) and (e).}
	\label{fig:circles}
\end{figure}

\section{Unraveling Time} 
\label{sec:ambiguities}
In our data-driven approach we define a dataset of \emph{input data} (a blurry image) and \emph{target} (a sequence of frames) pairs and then train a neural network to learn this mapping. However, the averaging of frames in model~\eqref{eq:forward} destroys the temporal ordering of the sequence $x[1],\dots, x[T]$. This makes the recovery of the frames $x[i]$ very challenging, because it is not possible to define the target uniquely. We might expect that local temporal ambiguities between subsequent frames can be resolved by learning the temporal smoothness (frames are more likely to form a sequence describing smooth motions). However, several other ambiguities  still remain. For example, the global motion direction is valid both forward and backward in time. This directional ambiguity applies independently to each moving object in the scene so that all motion direction combinations are valid.

We illustrate these ambiguities in Fig.~\ref{fig:circles} with a toy example. We consider two moving objects: a red and a green ball both translating along the horizontal axis. The first $5$ columns show all $5$ frames ($T=5$) in the averaging model. Because there are $2$ objects, there are $4$ possible combinations of motion directions ($2^n$ motions with $n$ the number of objects). These are shown in the $4$ rows of the figure. Column (f) shows that the corresponding average of the frames is the same identical motion-blurred image in all $4$ cases.
Therefore, any of the target frames across the $4$ rows is a valid one, and it would be unfeasible for the network to learn to predict a specific choice for just one of these $4$ cases. Indeed, as we show in the Experiments section, training a network to predict a single frame results in a network that predicts a blurry output that is the average of the possible choices. 
There is one exception to these ambiguities. The middle frame in an odd-numbered sequence does not change across the $4$ cases. This explains why prior methods \cite{Nah_2017_CVPR,debluringWild_2017_GCPR} could successfully train a neural network to predict the middle frame.

To address the temporal ordering ambiguities we introduce novel loss functions. In the next section we explore different options and show how we arrived at our proposed loss function. These cases are also discussed and evaluated in the Experiments section.


\section{From Image to Video}


Our training data has been obtained from a GoPro Hero 5 and features videos at 240 frames per second. To obtain blurry frames at a standard real-time video rates (30 frames per second) we thus need to average $8$ frames. However, as we have shown in our previous section we can avoid ambiguities in the estimation of the middle frame by using an odd number of frames, and hence we use $T=7$. However, our method can generalize to other choices of $T$.
We denote the neural network that predicts the frame $x[i]$ with $\phi_i$. Since the middle frame $x[4]$ can be predicted directly without ambiguities, we train $\phi_4$ with the following loss
\begin{equation}
{\cal L}_\text{middle} = | \phi_4(y) - x[4] |^2 + {\cal L}_\text{perceptual}(\phi_4(y),x[4]),
\label{eq:middle}
\end{equation}
where ${\cal L}_\text{perceptual}$ is the perceptual loss \cite{JohnsonAF16}. For the perceptual loss, we use the \textit{relu2\_2} and \textit{relu3\_3} layers of vgg16 net \cite{SimonyanZ14a}.
All other losses in the sections below will focus on the other frames.

\subsection{Globally Ordering-Invariant Loss}

A first way to recover all other frames is to use a loss function based on the image formation model~\eqref{eq:forward}
\begin{equation}
\textstyle{\cal L}_\text{model} = \left| \sum_{i\neq 4} \hat{x}[i] - \sum_{i\neq 4} x[i] \right|_1,
\label{eq:modelSolution}
\end{equation}
where we have defined $\hat{x}[i] = \phi_i(y)$.
This loss does not suffer from ambiguities and lets the networks decide what frames to output.
In practice, however, we find that it is too weak. This loss works well only when a blurry frame is generated by averaging no more than 3 frames. We find experimentally that with more averaging frames, the network does not converge well and may not generate a meaningful sequence.

\subsection{Pairwise Ordering-Invariant Loss}

Inspired by the previous observation, we notice that any pair of symmetric frames (about the middle frame) results in the same average and absolute differences.
This choice is motivated by the observations made in the previous section and illustrated in Fig.~\ref{fig:circles}. Columns (g) and (i) in Fig.~\ref{fig:circles} show the average and absolute difference respectively of columns (b) and (d).
These combinations yield the same target frame regardless of the objects motion direction.
Thus, we propose to use a loss made of two components, one based on the sum and the other based on the absolute difference between only two frames. We find experimentally that this scheme imposes a much stronger constraint. 
Based on these observations, for each pair of symmetric frames $(\phi_i, \phi_{8-i})$, we propose the following loss function
\begin{equation}
\begin{split}
\textstyle{\cal L}_\text{pair}= \sum_{i=1}^3 &\textstyle \Big| |\hat{x}[i] + \hat{x}[8-i]| - |x[i] + x[8-i]| \Big|_1 \\
 +&\textstyle\Big| |\hat{x}[i] - \hat{x}[8-i]| - |x[i] - x[8-i]| \Big|_1,
\label{eq:SumDiff_blurryInd}
\end{split}
\end{equation}
where $\hat{x}[i] = \phi_i(y)$ and $\hat{x}[8-i]=\phi_{8-i}(\phi_i(y), y)$ for $i=1,2,3$. Notice that $\phi_{8-i}(\phi_i(y), y)$ takes as inputs both the blurry image $y$ and the output of the other network $\phi_i(y)$. The reason for this additional input is so that the network $\phi_{8-i}$ can learn to generate a frame different from that of $\phi_i(y)$. Therefore, it needs to ``know'' what frame the network $\phi_i(y)$ has chosen to generate.
Compared with the loss function in eq.~\eqref{eq:modelSolution}, this loss function is easier to optimize and converges better (see results in the Experiments section). 
We also find experimentally that we can further boost the performance of our networks by additionally feeding the middle frame prediction to each network. That is, we define $\hat{x}[i]=\phi_{i}(\phi_4(y), y)$ and $\hat{x}[8-i] = \phi_{8-i}(\phi_4(y), \phi_i(y),y)$ for $i=1,2,3$.

\subsection{Learning a Temporal Direction}
\label{sec:training}

Up to this point, each pair of networks $\phi_{i}$ and $\phi_{8-i}$ operates independently from the other pairs. This is not ideal, as it leaves a binary ambiguity in the temporal ordering of each pair: We do not know if $\phi_{i}\mapsto x[i]$ and $\phi_{8-i}\mapsto x[8-i]$ or $\phi_{i}\mapsto x[8-i]$ and $\phi_{8-i}\mapsto x[i]$. Thus, after training, one needs to find a smooth temporal ordering of the outputs of these networks for each new input.

To avoid this additional task, we sequentially train our networks and use the outputs of the previous networks to determine the ordering for each data sample during training. This is needed only for frames away from the middle core with $i=3,4,5$. Moreover, once the temporal ordering chosen by the middle core networks is known to the other networks, there is no need to feed other inputs. Hence, we define the non-core networks as 
$\phi_{i}(\phi_{i+1}(y),\phi_{i+2}(y), y)$ and $\phi_{8-i}(\phi_{7-i}(y), \phi_{6-i}(y), y)$ for $i=1,2$. In practice, we find that $\phi_i$ and $\phi_{8-i}$ can share weights for $i=1,2$. This opens up the possibility of designing a recurrent network to predict all non-core frames. We also use an adversarial loss ${\cal L}_\text{adv}$ to enhance the accuracy of the output of each network $\phi$. Except for the network that generates the middle frame, all other networks use the adversarial loss during training.
We summarize our training losses and procedure in Table~\ref{tab:training}.

\begin{table}
\caption{Summary of networks, loss functions and training.\label{tab:training}}
\begin{tcolorbox}
Training procedure
\begin{enumerate}

\item Let $\hat{x}[4] = \phi_{4}(y)$ and minimize 
\begin{align}
{\cal L}_\text{middle} = |\hat{x}[4] - x[4]|^2+{\cal L}_\text{perceptual}(\hat{x}[4], x[4]).\nonumber
\end{align}

\item Let $\hat{x}[3] = \phi_{3}(\phi_4(y), y)$,\\ $\hat{x}[5] = \phi_{5}(\phi_3(y),\phi_4(y), y)$ and minimize 
\begin{align}
{\cal L}_\text{pair}^{3,5} = & \Big| |\hat{x}[3] + \hat{x}[5]| - |x[3] + x[5]| \Big|_1 \nonumber\\
 +&\Big| |\hat{x}[3] - \hat{x}[5]| - |x[3] - x[5]| \Big|_1\nonumber\\
 +&{\cal L}_\text{adv}^{3}+{\cal L}_\text{adv}^{5}.\nonumber
 \end{align}

\item Let $\hat{x}[i] = \phi_{i}(\phi_{i+1}(y),\phi_{i+2}(y), y)$, $\hat{x}[8-i] = \phi_{8-i}(\phi_{7-i}(y), \phi_{6-i}(y), y)$, with $i=1,2$ and minimize 
\begin{align}
{\cal L}_\text{pair}^{1,2,6,7} 
= & \Big| |\hat{x}[1] + \hat{x}[6]| - |x[1] + x[6]| \Big|_1\nonumber\\
 +& \Big| |\hat{x}[1] - \hat{x}[6]| - |x[1] - x[6]| \Big|_1\nonumber\\
 +& \Big| |\hat{x}[2] + \hat{x}[7]| - |x[2] + x[7]| \Big|_1\nonumber\\
 +& \Big| |\hat{x}[2] - \hat{x}[7]| - |x[2] - x[7]| \Big|_1\nonumber\\
 +& {\cal L}_\text{adv}^{1}+{\cal L}_\text{adv}^{2}+{\cal L}_\text{adv}^{6}+{\cal L}_\text{adv}^{7}.\nonumber
\end{align}

\end{enumerate}
\end{tcolorbox}
\end{table}


\section{Implementation} 
\label{sec:implementation}
Our middle frame prediction network employs a residual learning strategy like many recent image restoration networks \cite{Nah_2017_CVPR,Kim_2016_CVPR}. It consists of three parts, feature extraction, feature refinement and feature fusion.
Feature extraction is done by a resampling convolution with a resampling factor equal to 4.
In the feature refinement part, we use 12 residual blocks \cite{He_2016_CVPR}, where each one includes two $3\times 3$ and one $1\times 1$ convolution layers with a pre-activation structure \cite{He_2016_ECCV}. To further increase the receptive field, we replace standard convolutional layers in the middle with 6 residual blocks with dilated convolutions. The feature extraction and refinement parts work on grayscale images, and three color-refined features are generated separately. The feature fusion part works on color images to compensate misalignments from the three separately-generated color-refined features. For the non middle frame prediction networks, a similar structure is applied. The differences are the feature extraction part, where features are extracted from multiple inputs separately and then concatenated, resampling factor, and number of channels. More details of these architectures can be found in the supplementary material.

\section{Experiments}
\label{sec:experiments}
In this section, we perform a quantitative comparison of the middle frame prediction network with the state-of-the-art method \cite{Nah_2017_CVPR}. For non middle frame predictions, we carry out a qualitative evaluation as there is no existing method predicting a video sequence from a single motion-blurred input. We show some examples of video reconstructions from real motion blurred images in Fig.~\ref{fig:realResults}. More examples and videos are available in the supplementary material.
We validate our design through ablation studies of different loss functions.

\begin{table}[t]
\caption{Comparison of the middle frame prediction networks.}
\centering
\small
\begin{tabularx}{\linewidth}{@{}X c@{\hspace{1em}}c@{\hspace{1em}}c@{\hspace{1em}}c@{\hspace{1em}}c@{}}
\toprule
Method & $45\%$ \cite{Nah_2017_CVPR} (dB)   & Our testset  (dB) &  \cite{Nah_2017_CVPR} (dB)\\
\midrule
Nah \cite{Nah_2017_CVPR} &30.52  &28.19 & 28.48   \\
Middle   &32.20  &29.02  &26.98 \\
   \bottomrule
\end{tabularx}
\label{tab:MidFrameTab}
\vspace{-0.5em}
\end{table}

\begin{table}[t]
\caption{Execution time comparison between the state of the art single image dynamic scene deblurring network \cite{Nah_2017_CVPR} and our model on three different resolutions on a Titan X GPU.}
\centering
\small
\begin{tabularx}{\linewidth}{@{}X c@{\hspace{1em}}c@{\hspace{1em}}c@{\hspace{1em}}c@{\hspace{1em}}c@{}}
\toprule
Method &$320$P &$480$P   & $720$P & $\#$ params\\
\midrule
Nah \cite{Nah_2017_CVPR} &2.43  &3.52 &4.80  & 12M \\
Middle   &0.24  &0.30 &0.45 & 5M \\
Full   &0.61  &0.74 &1.10  & 17M \\
   \bottomrule
\end{tabularx}
\label{tab:Speed}
\vspace{-0.5em}
\end{table}
\begin{figure*}[t]
  \centering	
  	\rotatebox{90}{\hspace{.5cm} \small (a)}
	\begin{subfigure}[b]{0.156\textwidth}
	\includegraphics[width= \linewidth]{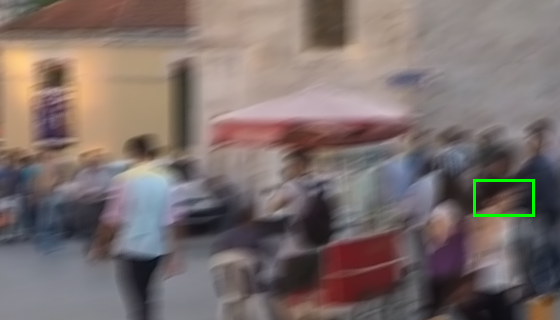}
	\end{subfigure}
	\begin{subfigure}[b]{0.156\textwidth}
	\includegraphics[width=\linewidth]{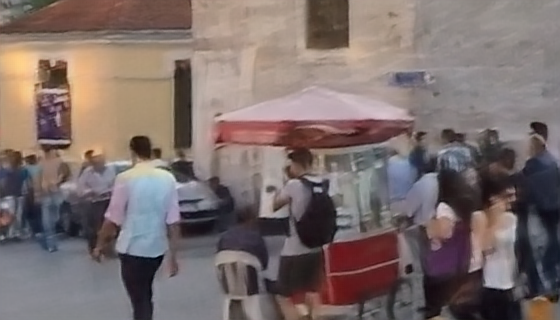}
	\end{subfigure}
	\begin{subfigure}[b]{0.156\textwidth}
	\includegraphics[width=\linewidth]{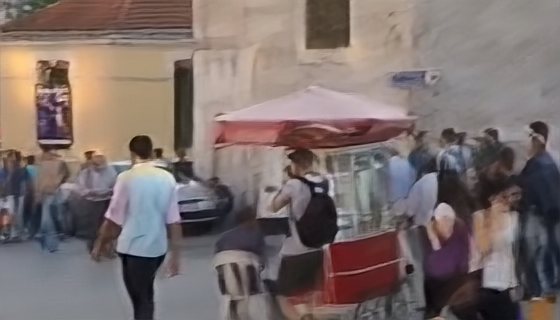}
	\end{subfigure}	
	\begin{subfigure}[b]{0.156\textwidth}
	\includegraphics[width= \linewidth]{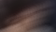}
	\end{subfigure}
	\begin{subfigure}[b]{0.156\textwidth}
	\includegraphics[width= \linewidth]{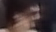}
	\end{subfigure}
	\begin{subfigure}[b]{0.156\textwidth}
	\includegraphics[width=\linewidth]{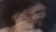}
	\end{subfigure}
  	\rotatebox{90}{\hspace{.5cm} \small (b)}
	\begin{subfigure}[b]{0.156\textwidth}
	\includegraphics[width= \linewidth]{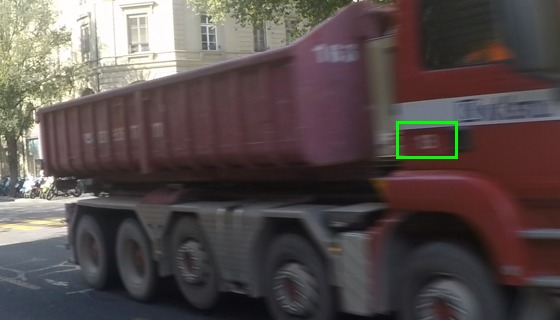}
	\end{subfigure}
	\begin{subfigure}[b]{0.156\textwidth}
	\includegraphics[width=\linewidth]{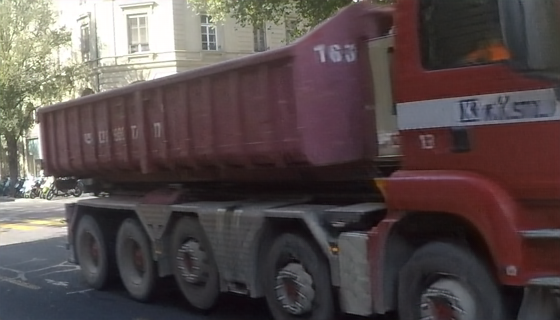}
	\end{subfigure}
	\begin{subfigure}[b]{0.156\textwidth}
	\includegraphics[width=\linewidth]{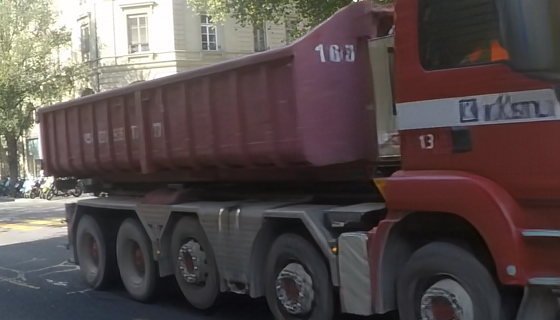}
	\end{subfigure}	
	\begin{subfigure}[b]{0.156\textwidth}
	\includegraphics[width= \linewidth]{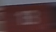}
	\end{subfigure}
	\begin{subfigure}[b]{0.156\textwidth}
	\includegraphics[width= \linewidth]{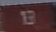}
	\end{subfigure}
	\begin{subfigure}[b]{0.156\textwidth}
	\includegraphics[width=\linewidth]{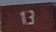}
	\end{subfigure}
  	\rotatebox{90}{\hspace{.5cm} \small (c)}
	\begin{subfigure}[b]{0.156\textwidth}
	\includegraphics[width= \linewidth]{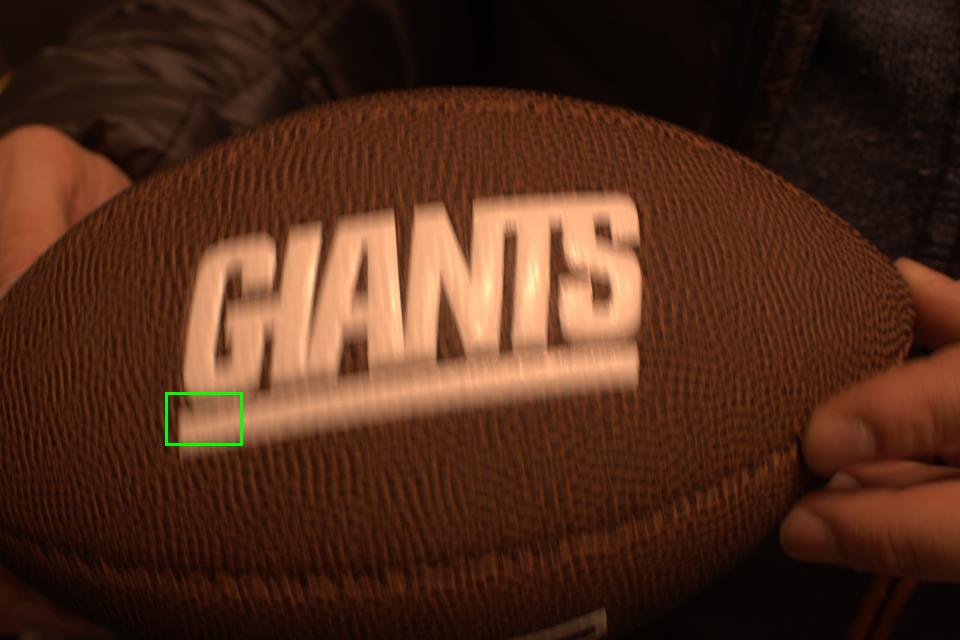}
	\end{subfigure}
	\begin{subfigure}[b]{0.156\textwidth}
	\includegraphics[width=\linewidth]{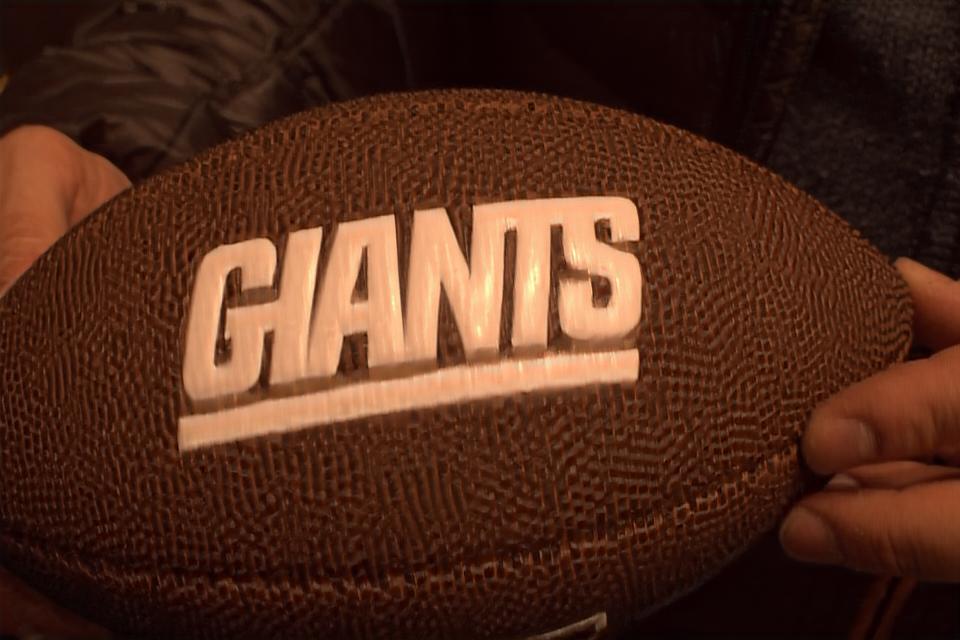}
	\end{subfigure}
	\begin{subfigure}[b]{0.156\textwidth}
	\includegraphics[width=\linewidth]{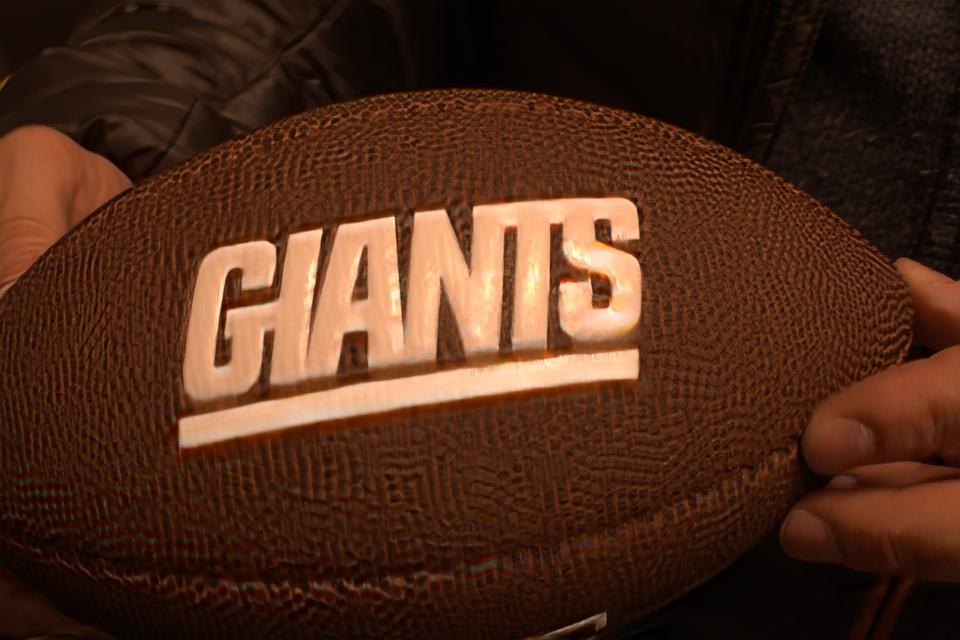}
	\end{subfigure}	
	\begin{subfigure}[b]{0.156\textwidth}
	\includegraphics[width= \linewidth]{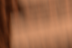}
	\end{subfigure}
	\begin{subfigure}[b]{0.156\textwidth}
	\includegraphics[width= \linewidth]{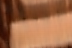}
	\end{subfigure}
	\begin{subfigure}[b]{0.156\textwidth}
	\includegraphics[width=\linewidth]{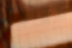}
	\end{subfigure}
	 \rotatebox{90}{\quad\quad\quad \small (d)}
	\begin{subfigure}[b]{0.156\textwidth}
	\includegraphics[width= \linewidth]{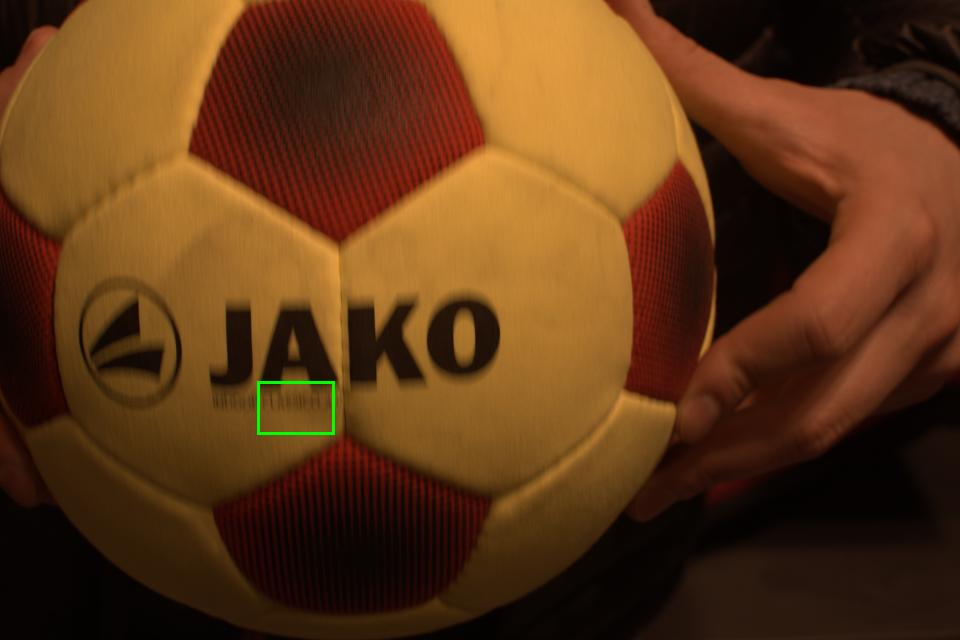}
	\caption*{Blurry}
	\end{subfigure}
	\begin{subfigure}[b]{0.156\textwidth}
	\includegraphics[width=\linewidth]{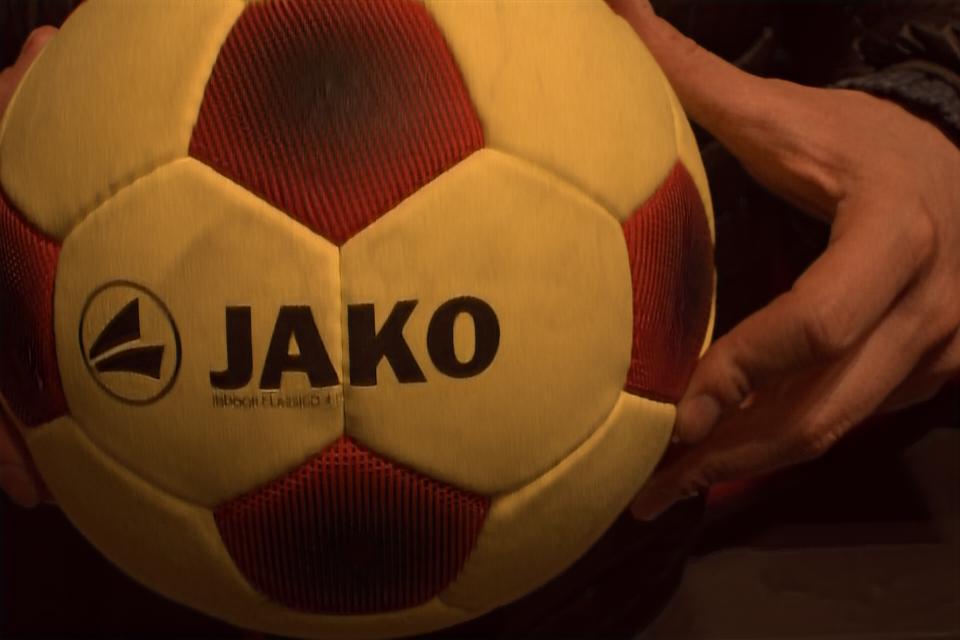}
	\caption*{Nah \cite{Nah_2017_CVPR}}
	\end{subfigure}
	\begin{subfigure}[b]{0.156\textwidth}
	\includegraphics[width=\linewidth]{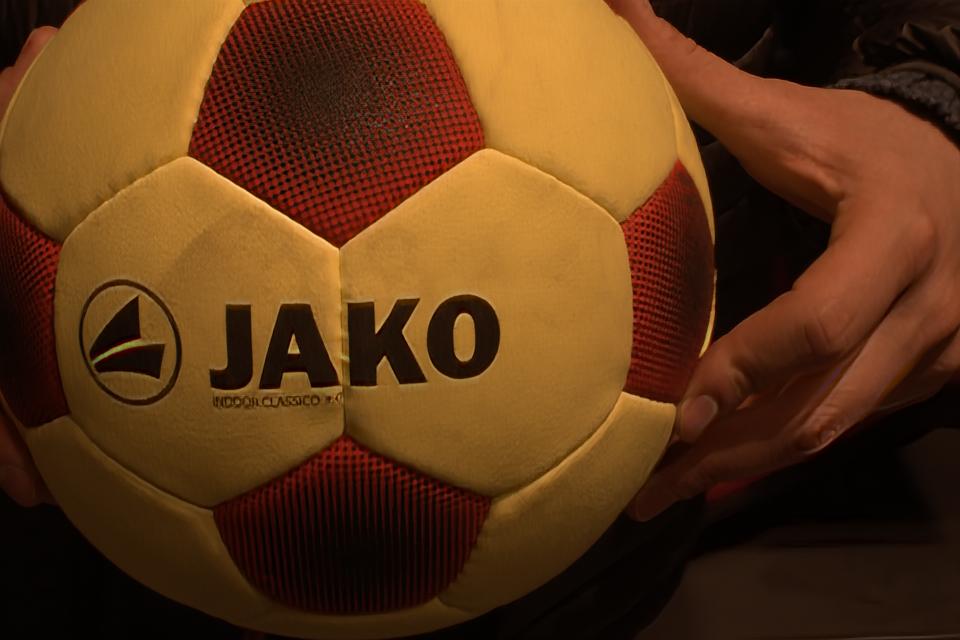}
	\caption*{Proposed}
	\end{subfigure}	
	\begin{subfigure}[b]{0.156\textwidth}
	\includegraphics[width= \linewidth]{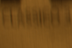}
	\caption*{Blurry  crop}
	\end{subfigure}
	\begin{subfigure}[b]{0.156\textwidth}
	\includegraphics[width= \linewidth]{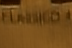}
	\caption*{Nah \cite{Nah_2017_CVPR} crop}
	\end{subfigure}
	\begin{subfigure}[b]{0.156\textwidth}
	\includegraphics[width=\linewidth]{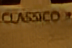}
	\caption*{Proposed crop}
	\end{subfigure}
  	\rotatebox{90}{\quad \small (e)}
	\begin{subfigure}[b]{0.156\textwidth}
	\includegraphics[width= \linewidth]{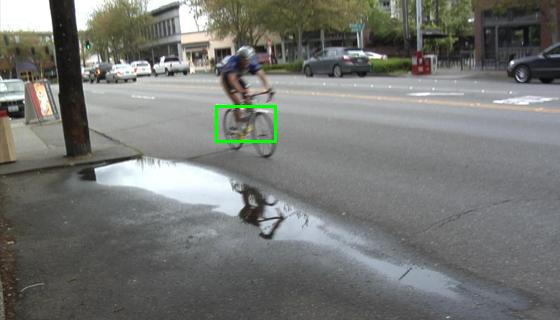}
	\end{subfigure}
	\begin{subfigure}[b]{0.156\textwidth}
	\includegraphics[width=\linewidth]{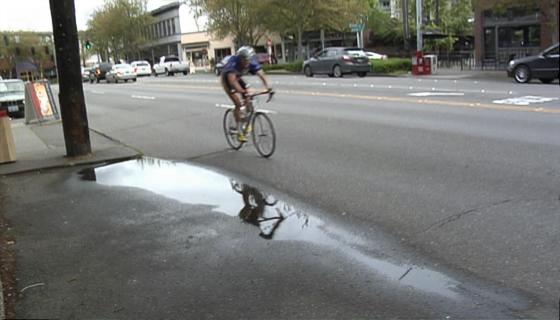}
	\end{subfigure}	
	\begin{subfigure}[b]{0.156\textwidth}
	\includegraphics[width= \linewidth]{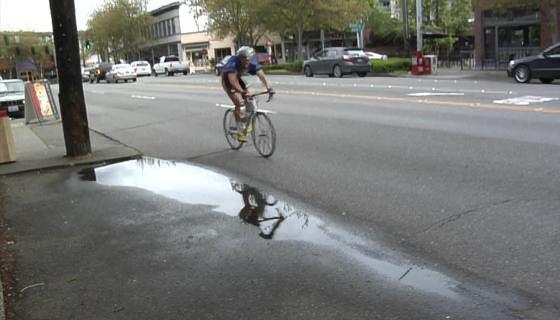}
	\end{subfigure}
	\begin{subfigure}[b]{0.156\textwidth}
	\includegraphics[width=\linewidth]{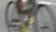}
	\end{subfigure}
	\begin{subfigure}[b]{0.156\textwidth}
	\includegraphics[width=\linewidth]{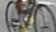}
	\end{subfigure}
	\begin{subfigure}[b]{0.156\textwidth}
	\includegraphics[width=\linewidth]{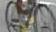}
	\end{subfigure}
  	\rotatebox{90}{\quad\quad \small (f)}
	\begin{subfigure}[b]{0.156\textwidth}
	\includegraphics[width= \linewidth]{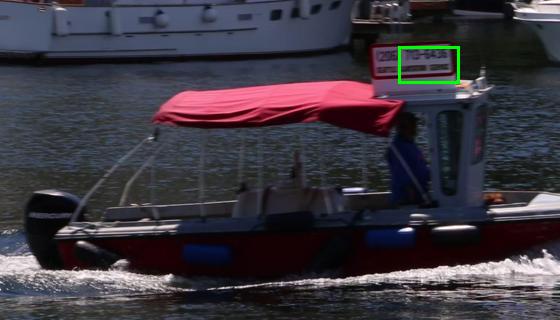}
	\end{subfigure}
	\begin{subfigure}[b]{0.156\textwidth}
	\includegraphics[width=\linewidth]{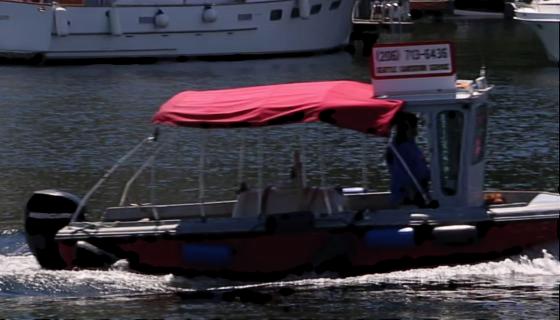}
	\end{subfigure}	
	\begin{subfigure}[b]{0.156\textwidth}
	\includegraphics[width= \linewidth]{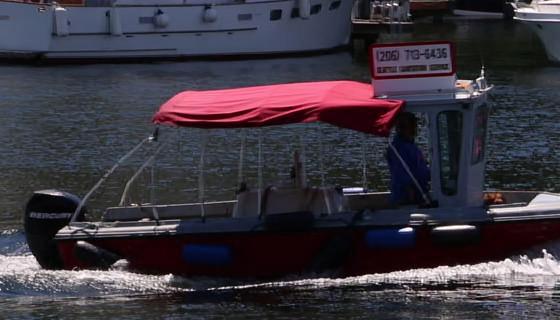}
	\end{subfigure}
	\begin{subfigure}[b]{0.156\textwidth}
	\includegraphics[width=\linewidth]{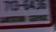}
	\end{subfigure}
	\begin{subfigure}[b]{0.156\textwidth}
	\includegraphics[width=\linewidth]{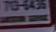}
	\end{subfigure}
	\begin{subfigure}[b]{0.156\textwidth}
	\includegraphics[width=\linewidth]{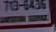}
	\end{subfigure}
  	\rotatebox{90}{\quad\quad\quad \small (g)}
	\begin{subfigure}[b]{0.156\textwidth}
	\includegraphics[width= \linewidth]{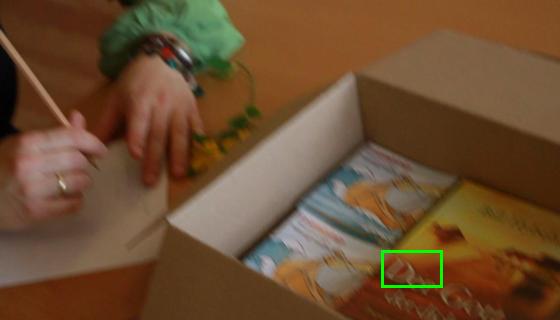}
	\caption*{Blurry}
	\end{subfigure}
	\begin{subfigure}[b]{0.156\textwidth}
	\includegraphics[width=\linewidth]{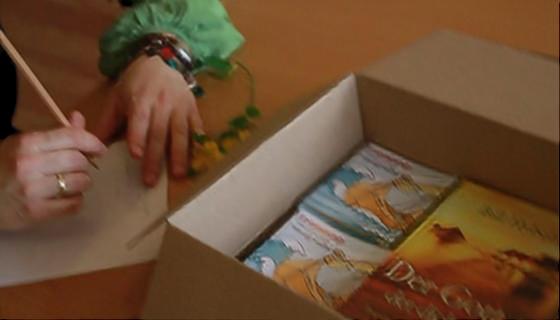}
	\caption*{Su \cite{Su_2017_CVPR}}
	\end{subfigure}	
	\begin{subfigure}[b]{0.156\textwidth}
	\includegraphics[width= \linewidth]{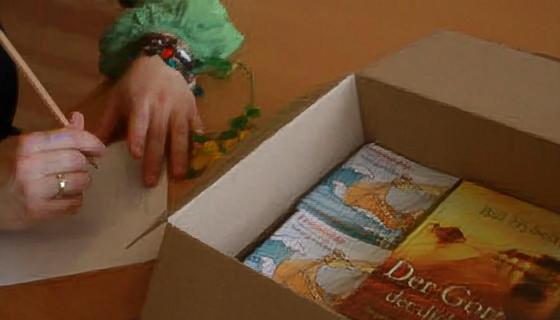}
	\caption*{Proposed}
	\end{subfigure}
	\begin{subfigure}[b]{0.156\textwidth}
	\includegraphics[width=\linewidth]{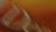}
	\caption*{Blurry crop}
	\end{subfigure}
	\begin{subfigure}[b]{0.156\textwidth}
	\includegraphics[width=\linewidth]{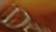}
	\caption*{Su \cite{Su_2017_CVPR} crop}
	\end{subfigure}
	\begin{subfigure}[b]{0.156\textwidth}
	\includegraphics[width=\linewidth]{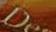}
	\caption*{Proposed crop}
	\end{subfigure}
	\caption{Middle frame prediction comparison. The first and fourth columns show the blurry inputs and cropped regions. (a)-(d) second and fifth columns: frame predictions from \cite{Nah_2017_CVPR} network. (e)-(g) second and fifth columns: frame predictions from \cite{Su_2017_CVPR} network. (a)-(g) third and last columns: frame predictions from our proposed network. The first two rows have been generated synthetically through averaging. The third and fourth rows are real images captured with a DSLR camera. The last three rows are real images from \cite{ChoWL12} and \cite{Su_2017_CVPR}.}
\label{fig:MiddleResults}
\vspace{-0.5em}
\end{figure*}

\noindent\textbf{Training Dataset and Implementation Details. }
 Although there is a GoPro training set available from \cite{Nah_2017_CVPR}, containing 22 diverse scenes, we captured additional 20 scenes. In the training, we downsample the GoPro frames to 45$\%$ of their original size ($1280\times 720$ pixels) to suppress noise. Blurry frames are generated by averaging $7$ consecutive frames randomly cropped of size $320\times 320$. For training we use about 15K samples. Data augmentation is applied to avoid overfitting by randomly shuffling color channels, rotating images, and adding $1\%$ white gaussian noise. Networks are implemented using PyTorch and training is done with 2 GTX 1080 Ti GPUs. The batch-size of the middle frame prediction network and other networks are 32 and 24, respectively. Training at each stage takes 1 day, and all together the full network training takes 4 days.

\begin{figure}[t]
  \centering
  	\hspace{.75em}
	\begin{subfigure}[b]{0.2235\textwidth}
	\includegraphics[width= \linewidth]{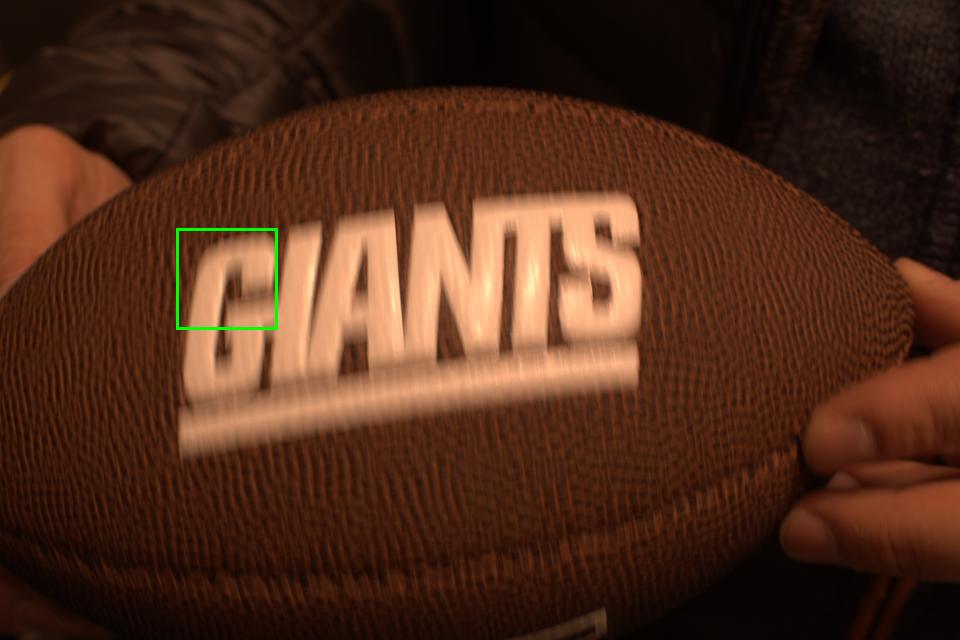}
	\caption*{Blurry}
	\end{subfigure}
	\begin{subfigure}[b]{0.2235\textwidth}
	\includegraphics[width= \linewidth]{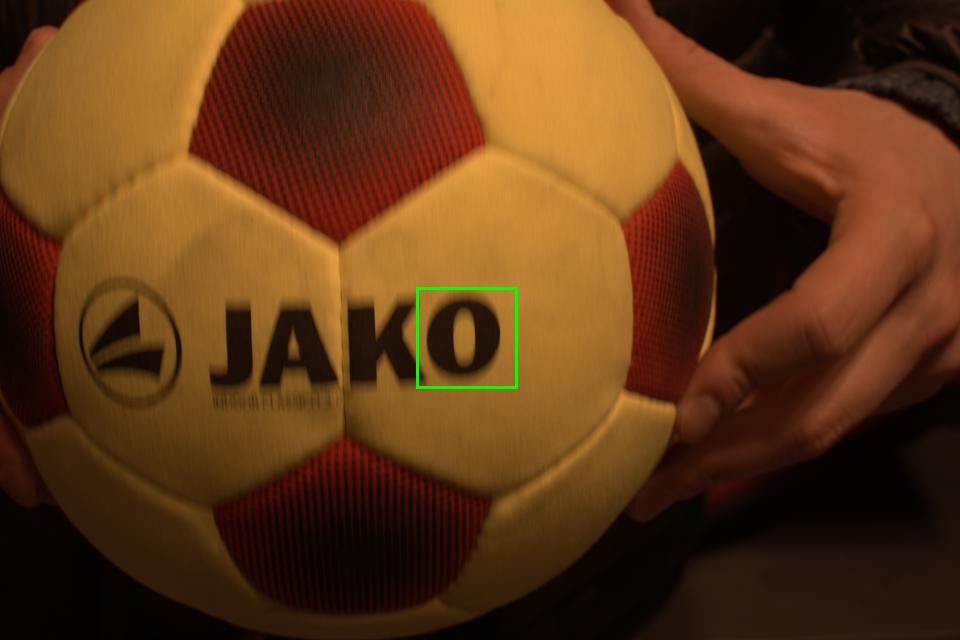}
	\caption*{Blurry}
	\end{subfigure}\\
	\rotatebox{90}{\hspace{.2cm} \small (a)}
	\begin{subfigure}[b]{0.052\textwidth}
	\includegraphics[width= \linewidth]{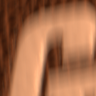}
	\end{subfigure}
	\begin{subfigure}[b]{0.052\textwidth}
	\includegraphics[width=\linewidth]{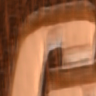}
	\end{subfigure}
	\begin{subfigure}[b]{0.052\textwidth}
	\includegraphics[width=\linewidth]{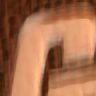}
	\end{subfigure}	
	\begin{subfigure}[b]{0.052\textwidth}
	\includegraphics[width= \linewidth]{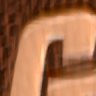}
	\end{subfigure}
	\begin{subfigure}[b]{0.052\textwidth}
	\includegraphics[width=\linewidth]{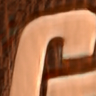}
	\end{subfigure}
	\begin{subfigure}[b]{0.052\textwidth}
	\includegraphics[width=\linewidth]{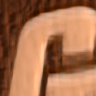}
	\end{subfigure}
	\begin{subfigure}[b]{0.052\textwidth}
	\includegraphics[width=\linewidth]{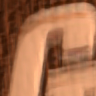}
	\end{subfigure}
	\begin{subfigure}[b]{0.052\textwidth}
	\includegraphics[width=\linewidth]{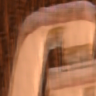}
	\end{subfigure}
	\rotatebox{90}{\hspace{.2cm} \small (b)}
	\begin{subfigure}[b]{0.052\textwidth}
	\includegraphics[width= \linewidth]{figures_DSLR_exp_7_image_38_blurry_patch}
	\end{subfigure}
	\begin{subfigure}[b]{0.052\textwidth}
	\includegraphics[width=\linewidth]{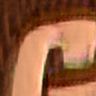}
	\end{subfigure}
	\begin{subfigure}[b]{0.052\textwidth}
	\includegraphics[width=\linewidth]{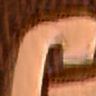}
	\end{subfigure}	
	\begin{subfigure}[b]{0.052\textwidth}
	\includegraphics[width= \linewidth]{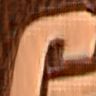}
	\end{subfigure}
	\begin{subfigure}[b]{0.052\textwidth}
	\includegraphics[width=\linewidth]{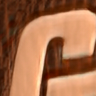}
	\end{subfigure}
	\begin{subfigure}[b]{0.052\textwidth}
	\includegraphics[width=\linewidth]{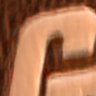}
	\end{subfigure}
	\begin{subfigure}[b]{0.052\textwidth}
	\includegraphics[width=\linewidth]{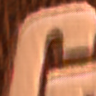}
	\end{subfigure}
	\begin{subfigure}[b]{0.052\textwidth}
	\includegraphics[width=\linewidth]{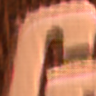}
	\end{subfigure}
	\rotatebox{90}{\hspace{.2cm}  \small (c)}
	\begin{subfigure}[b]{0.052\textwidth}
	\includegraphics[width= \linewidth]{figures_DSLR_exp_7_image_38_blurry_patch}
	\end{subfigure}
	\begin{subfigure}[b]{0.052\textwidth}
	\includegraphics[width=\linewidth]{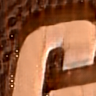}
	\end{subfigure}
	\begin{subfigure}[b]{0.052\textwidth}
	\includegraphics[width=\linewidth]{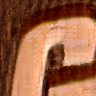}
	\end{subfigure}	
	\begin{subfigure}[b]{0.052\textwidth}
	\includegraphics[width= \linewidth]{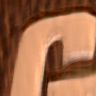}
	\end{subfigure}
	\begin{subfigure}[b]{0.052\textwidth}
	\includegraphics[width=\linewidth]{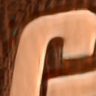}
	\end{subfigure}
	\begin{subfigure}[b]{0.052\textwidth}
	\includegraphics[width=\linewidth]{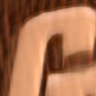}
	\end{subfigure}
	\begin{subfigure}[b]{0.052\textwidth}
	\includegraphics[width=\linewidth]{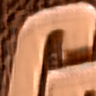}
	\end{subfigure}
	\begin{subfigure}[b]{0.052\textwidth}
	\includegraphics[width=\linewidth]{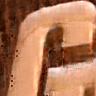}
	\end{subfigure}
	\rotatebox{90}{\hspace{.2cm}  \small (d)}
	\begin{subfigure}[b]{0.052\textwidth}
	\includegraphics[width= \linewidth]{figures_DSLR_exp_7_image_38_blurry_patch}
	\end{subfigure}
	\begin{subfigure}[b]{0.052\textwidth}
	\includegraphics[width=\linewidth]{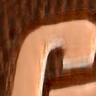}
	\end{subfigure}
	\begin{subfigure}[b]{0.052\textwidth}
	\includegraphics[width=\linewidth]{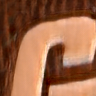}
	\end{subfigure}	
	\begin{subfigure}[b]{0.052\textwidth}
	\includegraphics[width= \linewidth]{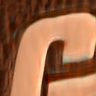}
	\end{subfigure}
	\begin{subfigure}[b]{0.052\textwidth}
	\includegraphics[width=\linewidth]{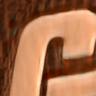}
	\end{subfigure}
	\begin{subfigure}[b]{0.052\textwidth}
	\includegraphics[width=\linewidth]{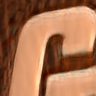}
	\end{subfigure}
	\begin{subfigure}[b]{0.052\textwidth}
	\includegraphics[width=\linewidth]{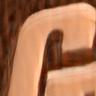}
	\end{subfigure}
	\begin{subfigure}[b]{0.052\textwidth}
	\includegraphics[width=\linewidth]{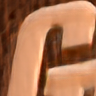}
	\end{subfigure}
	\rotatebox{90}{\hspace{.2cm}  \small (e)}		
	\begin{subfigure}[b]{0.052\textwidth}
	\includegraphics[width= \linewidth]{figures_DSLR_exp_7_image_38_blurry_patch}
	\end{subfigure}
	\begin{subfigure}[b]{0.052\textwidth}
	\includegraphics[width=\linewidth]{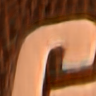}
	\end{subfigure}
	\begin{subfigure}[b]{0.052\textwidth}
	\includegraphics[width=\linewidth]{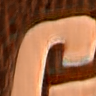}
	\end{subfigure}	
	\begin{subfigure}[b]{0.052\textwidth}
	\includegraphics[width= \linewidth]{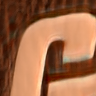}
	\end{subfigure}
	\begin{subfigure}[b]{0.052\textwidth}
	\includegraphics[width=\linewidth]{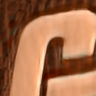}
	\end{subfigure}
	\begin{subfigure}[b]{0.052\textwidth}
	\includegraphics[width=\linewidth]{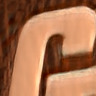}
	\end{subfigure}
	\begin{subfigure}[b]{0.052\textwidth}
	\includegraphics[width=\linewidth]{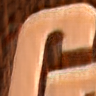}
	\end{subfigure}
	\begin{subfigure}[b]{0.052\textwidth}
	\includegraphics[width=\linewidth]{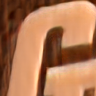}
	\end{subfigure}
	\rotatebox{90}{\hspace{.2cm}  \small (f)}		
	\begin{subfigure}[b]{0.052\textwidth}
	\includegraphics[width= \linewidth]{figures_DSLR_exp_7_image_38_blurry_patch}
	\end{subfigure}
	\begin{subfigure}[b]{0.052\textwidth}
	\includegraphics[width=\linewidth]{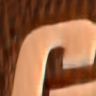}
	\end{subfigure}
	\begin{subfigure}[b]{0.052\textwidth}
	\includegraphics[width=\linewidth]{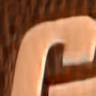}
	\end{subfigure}	
	\begin{subfigure}[b]{0.052\textwidth}
	\includegraphics[width= \linewidth]{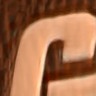}
	\end{subfigure}
	\begin{subfigure}[b]{0.052\textwidth}
	\includegraphics[width=\linewidth]{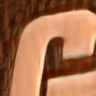}
	\end{subfigure}
	\begin{subfigure}[b]{0.052\textwidth}
	\includegraphics[width=\linewidth]{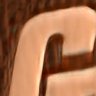}
	\end{subfigure}
	\begin{subfigure}[b]{0.052\textwidth}
	\includegraphics[width=\linewidth]{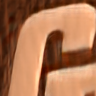}
	\end{subfigure}
	\begin{subfigure}[b]{0.052\textwidth}
	\includegraphics[width=\linewidth]{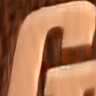}
	\end{subfigure}
	\rotatebox{90}{\hspace{.2cm}  \small (g)}
	\begin{subfigure}[b]{0.052\textwidth}
	\includegraphics[width= \linewidth]{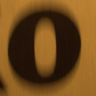}
	\end{subfigure}
	\begin{subfigure}[b]{0.052\textwidth}
	\includegraphics[width=\linewidth]{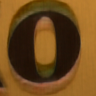}
	\end{subfigure}
	\begin{subfigure}[b]{0.052\textwidth}
	\includegraphics[width=\linewidth]{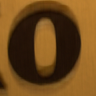}
	\end{subfigure}	
	\begin{subfigure}[b]{0.052\textwidth}
	\includegraphics[width= \linewidth]{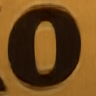}
	\end{subfigure}
	\begin{subfigure}[b]{0.052\textwidth}
	\includegraphics[width=\linewidth]{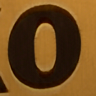}
	\end{subfigure}
	\begin{subfigure}[b]{0.052\textwidth}
	\includegraphics[width=\linewidth]{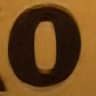}
	\end{subfigure}
	\begin{subfigure}[b]{0.052\textwidth}
	\includegraphics[width=\linewidth]{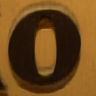}
	\end{subfigure}
	\begin{subfigure}[b]{0.052\textwidth}
	\includegraphics[width=\linewidth]{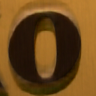}
	\end{subfigure}
	\rotatebox{90}{\hspace{.2cm}  \small (h)}
	\begin{subfigure}[b]{0.052\textwidth}
	\includegraphics[width= \linewidth]{figures_DSLR_exp_7_image_58_blurry_patch}
	\end{subfigure}
	\begin{subfigure}[b]{0.052\textwidth}
	\includegraphics[width=\linewidth]{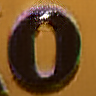}
	\end{subfigure}
	\begin{subfigure}[b]{0.052\textwidth}
	\includegraphics[width=\linewidth]{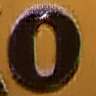}
	\end{subfigure}	
	\begin{subfigure}[b]{0.052\textwidth}
	\includegraphics[width= \linewidth]{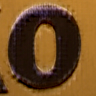}
	\end{subfigure}
	\begin{subfigure}[b]{0.052\textwidth}
	\includegraphics[width=\linewidth]{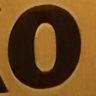}
	\end{subfigure}
	\begin{subfigure}[b]{0.052\textwidth}
	\includegraphics[width=\linewidth]{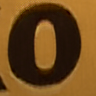}
	\end{subfigure}
	\begin{subfigure}[b]{0.052\textwidth}
	\includegraphics[width=\linewidth]{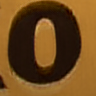}
	\end{subfigure}
	\begin{subfigure}[b]{0.052\textwidth}
	\includegraphics[width=\linewidth]{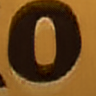}
	\end{subfigure}
	\rotatebox{90}{\hspace{.2cm}  \small (i)}
	\begin{subfigure}[b]{0.052\textwidth}
	\includegraphics[width= \linewidth]{figures_DSLR_exp_7_image_58_blurry_patch}
	\end{subfigure}
	\begin{subfigure}[b]{0.052\textwidth}
	\includegraphics[width=\linewidth]{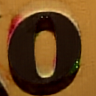}
	\end{subfigure}
	\begin{subfigure}[b]{0.052\textwidth}
	\includegraphics[width=\linewidth]{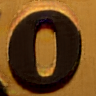}
	\end{subfigure}	
	\begin{subfigure}[b]{0.052\textwidth}
	\includegraphics[width= \linewidth]{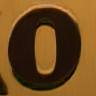}
	\end{subfigure}
	\begin{subfigure}[b]{0.052\textwidth}
	\includegraphics[width=\linewidth]{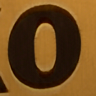}
	\end{subfigure}
	\begin{subfigure}[b]{0.052\textwidth}
	\includegraphics[width=\linewidth]{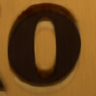}
	\end{subfigure}
	\begin{subfigure}[b]{0.052\textwidth}
	\includegraphics[width=\linewidth]{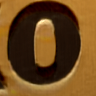}
	\end{subfigure}
	\begin{subfigure}[b]{0.052\textwidth}
	\includegraphics[width=\linewidth]{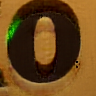}
	\end{subfigure}
	\rotatebox{90}{\hspace{.2cm}  \small (j)}
	\begin{subfigure}[b]{0.052\textwidth}
	\includegraphics[width= \linewidth]{figures_DSLR_exp_7_image_58_blurry_patch}
	\end{subfigure}
	\begin{subfigure}[b]{0.052\textwidth}
	\includegraphics[width=\linewidth]{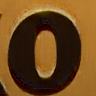}
	\end{subfigure}
	\begin{subfigure}[b]{0.052\textwidth}
	\includegraphics[width=\linewidth]{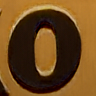}
	\end{subfigure}	
	\begin{subfigure}[b]{0.052\textwidth}
	\includegraphics[width= \linewidth]{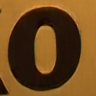}
	\end{subfigure}
	\begin{subfigure}[b]{0.052\textwidth}
	\includegraphics[width=\linewidth]{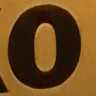}
	\end{subfigure}
	\begin{subfigure}[b]{0.052\textwidth}
	\includegraphics[width=\linewidth]{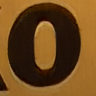}
	\end{subfigure}
	\begin{subfigure}[b]{0.052\textwidth}
	\includegraphics[width=\linewidth]{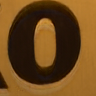}
	\end{subfigure}
	\begin{subfigure}[b]{0.052\textwidth}
	\includegraphics[width=\linewidth]{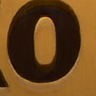}
	\end{subfigure}	
	\rotatebox{90}{\hspace{.2cm} \small (k)}		
	\begin{subfigure}[b]{0.052\textwidth}
	\includegraphics[width= \linewidth]{figures_DSLR_exp_7_image_58_blurry_patch}
	\end{subfigure}
	\begin{subfigure}[b]{0.052\textwidth}
	\includegraphics[width=\linewidth]{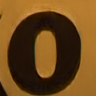}
	\end{subfigure}
	\begin{subfigure}[b]{0.052\textwidth}
	\includegraphics[width=\linewidth]{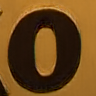}
	\end{subfigure}	
	\begin{subfigure}[b]{0.052\textwidth}
	\includegraphics[width= \linewidth]{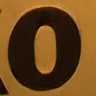}
	\end{subfigure}
	\begin{subfigure}[b]{0.052\textwidth}
	\includegraphics[width=\linewidth]{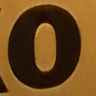}
	\end{subfigure}
	\begin{subfigure}[b]{0.052\textwidth}
	\includegraphics[width=\linewidth]{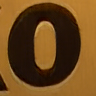}
	\end{subfigure}
	\begin{subfigure}[b]{0.052\textwidth}
	\includegraphics[width=\linewidth]{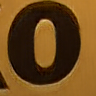}
	\end{subfigure}
	\begin{subfigure}[b]{0.052\textwidth}
	\includegraphics[width=\linewidth]{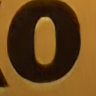}
	\end{subfigure}
	\rotatebox{90}{\hspace{.55cm}  \small (l)}
	\begin{subfigure}[b]{0.052\textwidth}
	\includegraphics[width= \linewidth]{figures_DSLR_exp_7_image_58_blurry_patch}
	\caption*{Blurry}
	\end{subfigure}
	\begin{subfigure}[b]{0.052\textwidth}
	\includegraphics[width=\linewidth]{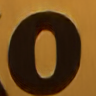}
	\caption*{Frame 1}
	\end{subfigure}
	\begin{subfigure}[b]{0.052\textwidth}
	\includegraphics[width=\linewidth]{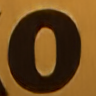}
	\caption*{Frame 2}
	\end{subfigure}	
	\begin{subfigure}[b]{0.052\textwidth}
	\includegraphics[width= \linewidth]{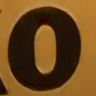}
	\caption*{Frame 3}
	\end{subfigure}
	\begin{subfigure}[b]{0.052\textwidth}
	\includegraphics[width=\linewidth]{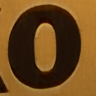}
	\caption*{Frame 4}
	\end{subfigure}
	\begin{subfigure}[b]{0.052\textwidth}
	\includegraphics[width=\linewidth]{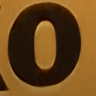}
	\caption*{Frame 5}
	\end{subfigure}
	\begin{subfigure}[b]{0.052\textwidth}
	\includegraphics[width=\linewidth]{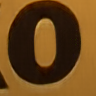}
	\caption*{Frame 6}
	\end{subfigure}
	\begin{subfigure}[b]{0.052\textwidth}
	\includegraphics[width=\linewidth]{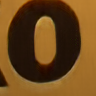}
	\caption*{Frame 7}
	\end{subfigure}
	\caption{Ablation study on real data with different loss functions.}
\label{fig:SchemeResults1}
\vspace{-1.5em}
\end{figure}

\begin{figure}[t]
  \centering	
    	\rotatebox{90}{\quad\quad\quad\quad\quad \small (a)}
	\begin{subfigure}[b]{0.225\textwidth}
	\includegraphics[width= \linewidth]{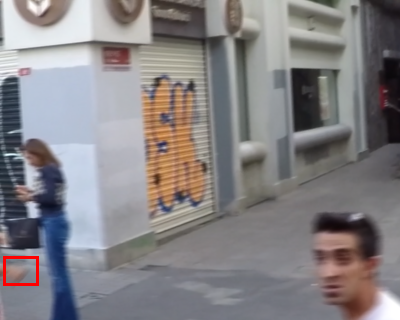}
	\caption*{Blurry}
	\end{subfigure}
	\begin{subfigure}[b]{0.225\textwidth}
	\includegraphics[width= \linewidth]{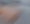}
	\caption*{Blurry crop}
	\end{subfigure}
    	\rotatebox{90}{\hspace{.75cm}  \small (b)}
	\begin{subfigure}[b]{0.148\textwidth}
	\includegraphics[width=\linewidth]{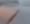}
	\end{subfigure}
	\begin{subfigure}[b]{0.148\textwidth}
	\includegraphics[width=\linewidth]{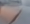}
	\end{subfigure}
	\begin{subfigure}[b]{0.148\textwidth}
	\includegraphics[width=\linewidth]{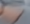}
	\end{subfigure}
    	\rotatebox{90}{\hspace{.75cm} \small (c)}
	\begin{subfigure}[b]{0.148\textwidth}
	\includegraphics[width=\linewidth]{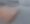}
	\end{subfigure}
	\begin{subfigure}[b]{0.148\textwidth}
	\includegraphics[width=\linewidth]{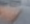}
	\end{subfigure}
	\begin{subfigure}[b]{0.148\textwidth}
	\includegraphics[width=\linewidth]{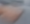}
	\end{subfigure}
    	\rotatebox{90}{\quad\quad\quad\quad \small (d)}
	\begin{subfigure}[b]{0.148\textwidth}
	\includegraphics[width=\linewidth]{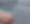}
	\caption*{Frame 1}
	\end{subfigure}
	\begin{subfigure}[b]{0.148\textwidth}
	\includegraphics[width=\linewidth]{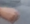}
	\caption*{Frame 4}
	\end{subfigure}
	\begin{subfigure}[b]{0.148\textwidth}
	\includegraphics[width=\linewidth]{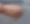}
	\caption*{Frame 7}
	\end{subfigure}
	\caption{A synthetic example from \cite{Nah_2017_CVPR} test image. (b)-(d) Frame 4 show 
our estimated middle frame, Nah's estimate and the ground truth, respectively. As can be seen, the middle frame estimates from both our method and Nah's are incorrect and affect the estimates of the other frames. Only when the ground truth middle frame is provided, the other frames can be estimated correctly.}
\label{fig:FailResults}
\vspace{-1.0em}
\end{figure}
\noindent\textbf{Middle Frame Reconstruction. }
We take Nah's \cite{Nah_2017_CVPR} test set, which contains 11 different sequences, and generate 1700 blurry frames by averaging 7 consecutive frames. The same process is also applied to our own test set where 450 blurry images are generated. All blurry images are downsampled to $45\%$ as during training.
Table~\ref{tab:MidFrameTab} shows the quantitative results of Nah's network and our proposed network on two datasets. On the last two datasets (the first two columns in the table), our network is consistently preforming better. This is because the motion blur in the data matches the motion blur observed by our network during training. In contrast, Nah's network was trained with much more challenging data, where motion blur could be even larger. Thus, for a fairer comparison, we also evaluate our network on Nah's original $1111$ test images. These images are averaged by more than $7$ frames without any downsampling. In this case, Nah's network is performing  better, as our network has not learned to deal with such large motion blur. However, the performance loss is not too significant.

Some visual comparisons on both synthetic and real images are shown in Fig.~\ref{fig:MiddleResults}. Fig.~\ref{fig:MiddleResults} (a) and (b) show two synthetic examples, one with an extremely large blur from Nah's original test images and the other one with a moderate blur from our test set. It can be seen that although our method does not outperform Nah's, it can give better visual results when blur is moderate. Two real examples captured with a DSLR (Nikon D7100) are also shown in rows (c) and (d). 
In practice, we find that if a network is trained with large blurs, it may not remove moderate blur to the same extent as networks trained with small blurs. As we will show later in the Experiments section, the accuracy of the middle frame prediction has a dramatic impact on the reconstruction of the other frames. 

Another advantage of our network is computational efficiency. In Table~\ref{tab:Speed} we show the execution time for 3 different resolutions and number of parameters used in Nah's and our networks. It can be seen that our middle frame prediction network is approximately 10 times faster than Nah's, but has half as many parameters. Additionally, in Fig.~\ref{fig:MiddleResults} we also compare to the state of the art video deblurring method \cite{Su_2017_CVPR}. Notice that in \cite{Su_2017_CVPR}, they use 5 consecutive blurry frames to predict the middle sharp frame, whereas we predict the middle frame directly from only one blurry input. Three real results are shown in Fig.~\ref{fig:MiddleResults} (e), (f) and (g). It can be seen that, although our method suffers from some jpeg artifacts, it gives comparable results.

\noindent\textbf{Resampling Factors. }
We also evaluate the different resampling factors for the middle frame estimation and find that $4\times$ resampling gives a better trade-off between accuracy and execution time.

\begin{figure*}[t]
  \centering
\includegraphics[width=0.8\linewidth,trim={0 7cm 0 0},clip]{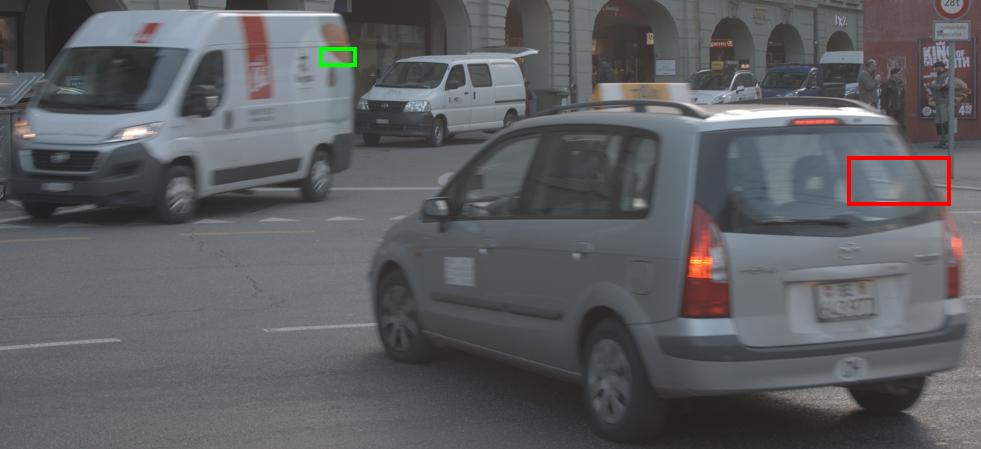}\\%
	\includegraphics[width=0.124\linewidth,trim={0 7cm 0 0},clip]{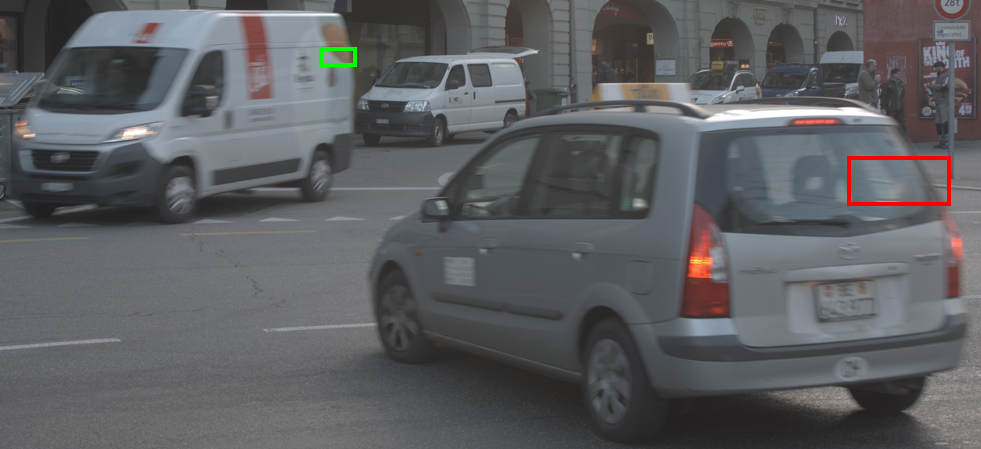}
	\includegraphics[width=0.124\linewidth,trim={0 7cm 0 0},clip]{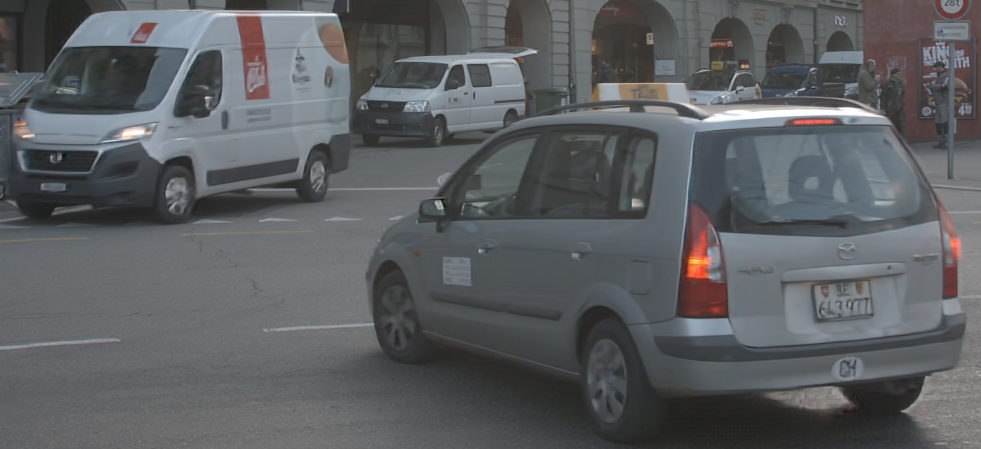}%
	\includegraphics[width=0.124\linewidth,trim={0 7cm 0 0},clip]{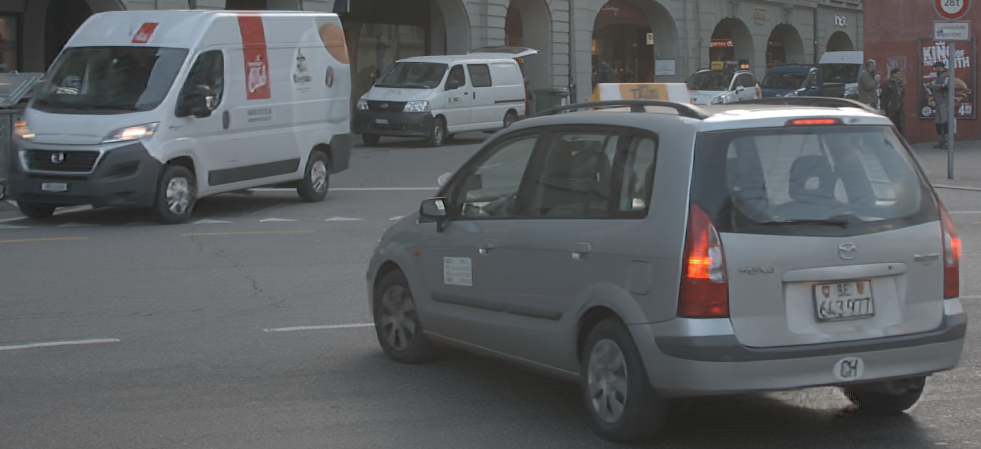}%
	\includegraphics[width=0.124\linewidth,trim={0 7cm 0 0},clip]{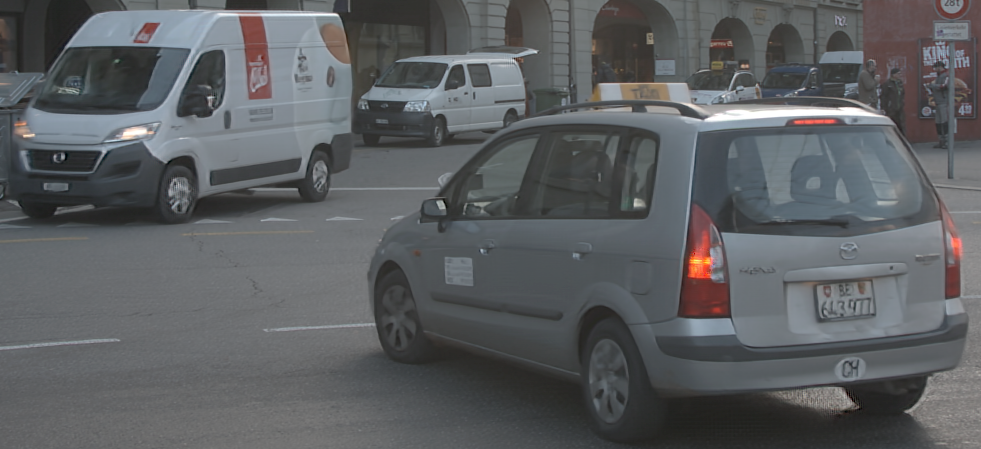}%
	\includegraphics[width=0.124\linewidth,trim={0 7cm 0 0},clip]{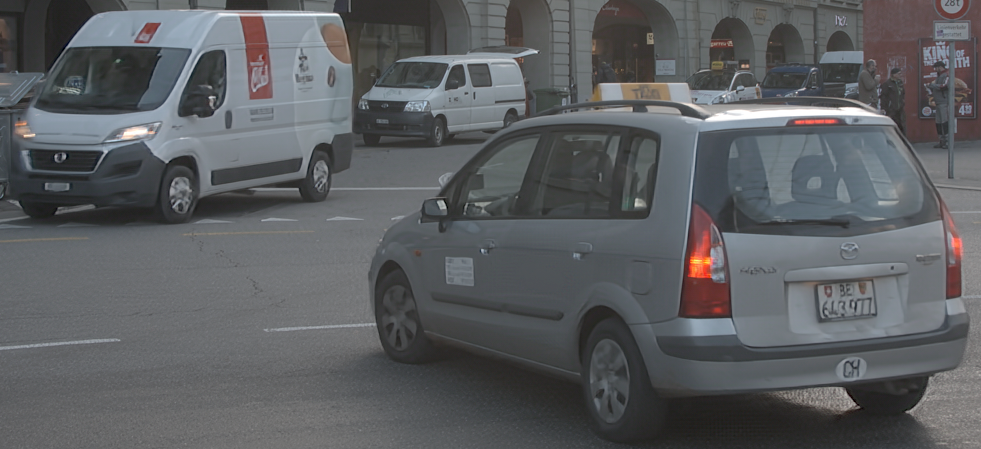}%
	\includegraphics[width=0.124\linewidth,trim={0 7cm 0 0},clip]{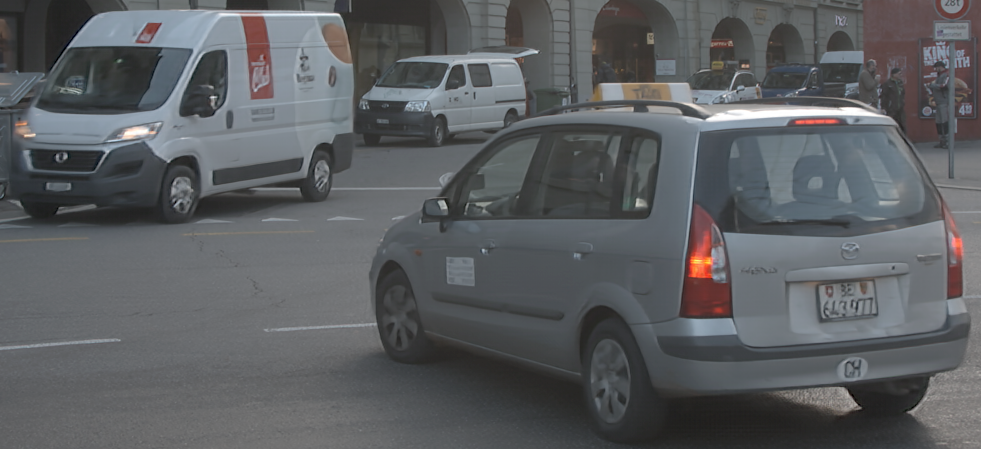}%
	\includegraphics[width=0.124\linewidth,trim={0 7cm 0 0},clip]{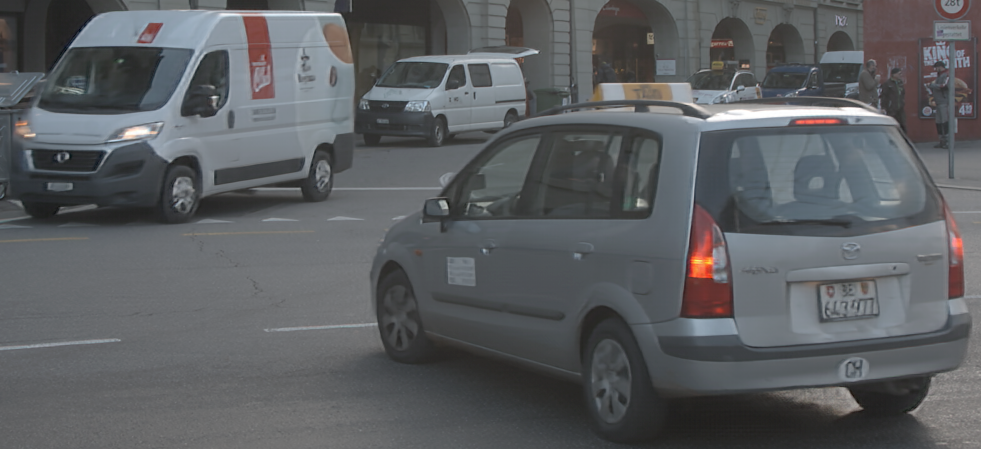}%
	\includegraphics[width=0.124\linewidth,trim={0 7cm 0 0},clip]{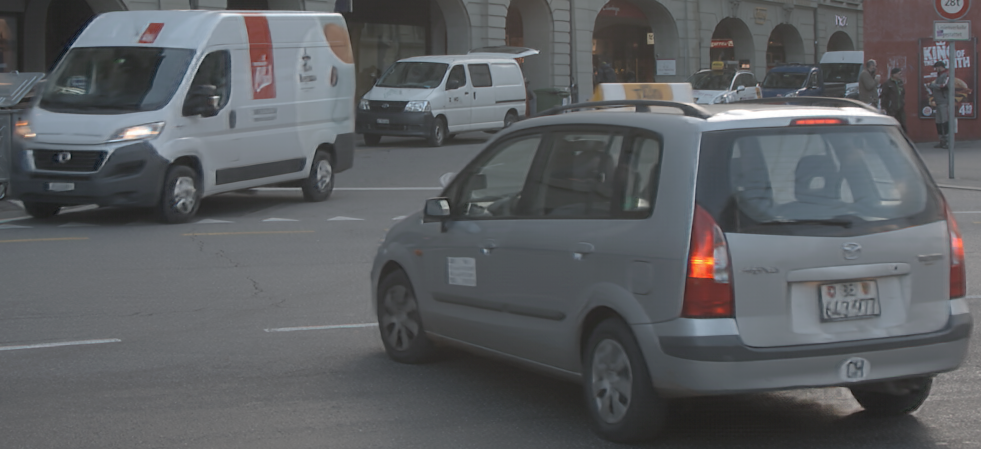}\\%
	\includegraphics[width=0.124\linewidth]{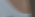}
	\includegraphics[width=0.124\linewidth]{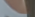}%
	\includegraphics[width=0.124\linewidth]{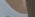}%
	\includegraphics[width=0.124\linewidth]{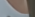}%
	\includegraphics[width=0.124\linewidth]{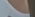}%
	\includegraphics[width=0.124\linewidth]{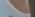}%
	\includegraphics[width=0.124\linewidth]{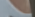}%
	\includegraphics[width=0.124\linewidth]{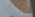}\\
	\includegraphics[width=0.124\linewidth]{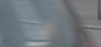}
	\includegraphics[width=0.124\linewidth]{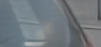}%
	\includegraphics[width=0.124\linewidth]{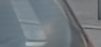}%
	\includegraphics[width=0.124\linewidth]{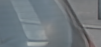}%
	\includegraphics[width=0.124\linewidth]{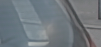}%
	\includegraphics[width=0.124\linewidth]{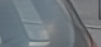}%
	\includegraphics[width=0.124\linewidth]{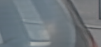}%
	\includegraphics[width=0.124\linewidth]{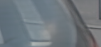}\\
	\begin{subfigure}[b]{0.124\textwidth}
  	\includegraphics[width=\linewidth,trim={0cm 0cm 7cm 0cm},clip]{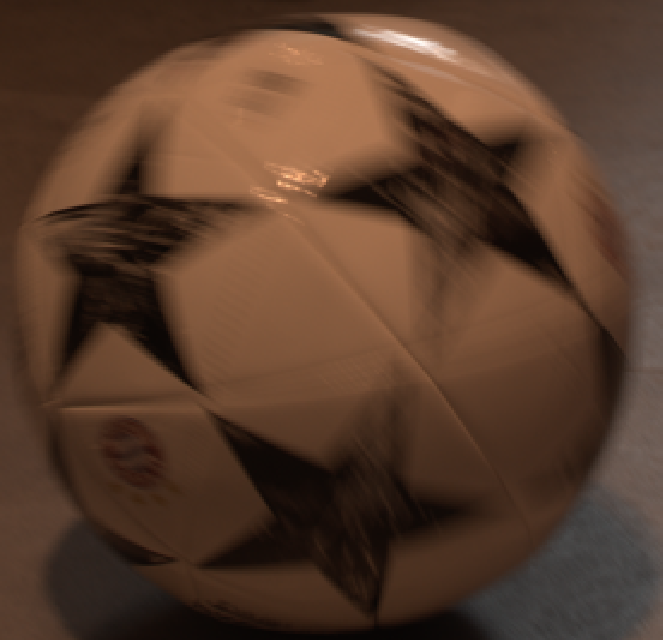}
	\caption*{Blurry}
	\end{subfigure}
	\begin{subfigure}[b]{0.124\textwidth}
	\includegraphics[width=\linewidth,trim={0cm 0cm 7cm 0cm},clip]{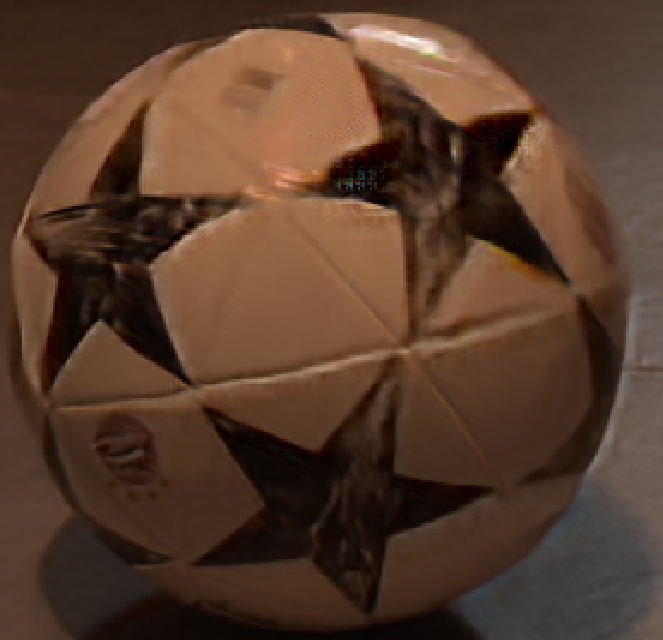}%
	\caption*{Frame 1}
	\end{subfigure}%
	\begin{subfigure}[b]{0.124\textwidth}
	\includegraphics[width=\linewidth,trim={0cm 0cm 7cm 0cm},clip]{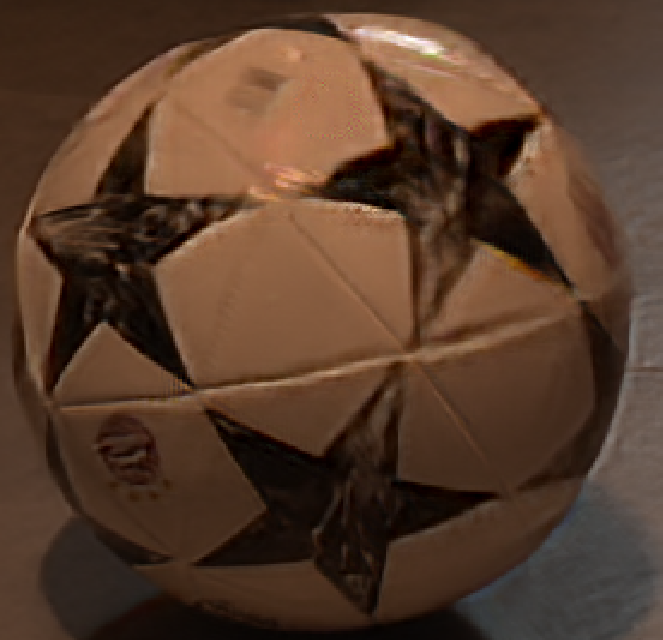}%
	\caption*{Frame 2}
	\end{subfigure}%
	\begin{subfigure}[b]{0.124\textwidth}
	\includegraphics[width=\linewidth,trim={0cm 0cm 7cm 0cm},clip]{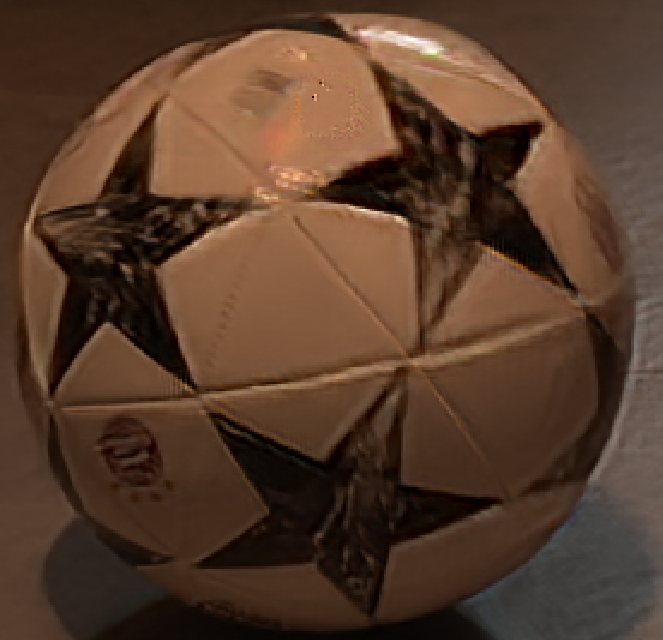}%
	\caption*{Frame 3}
	\end{subfigure}%
	\begin{subfigure}[b]{0.124\textwidth}
	\includegraphics[width=\linewidth,trim={0cm 0cm 7cm 0cm},clip]{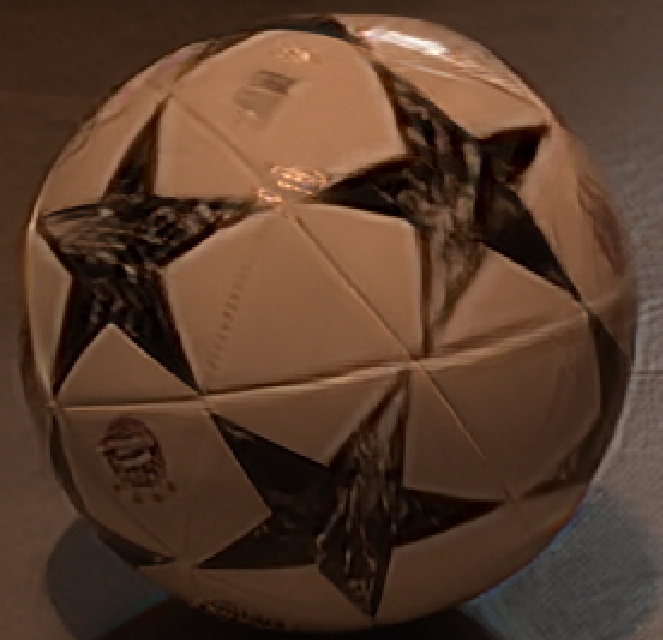}%
	\caption*{Frame 4}
	\end{subfigure}%
	\begin{subfigure}[b]{0.124\textwidth}
	\includegraphics[width=\linewidth,trim={0cm 0cm 7cm 0cm},clip]{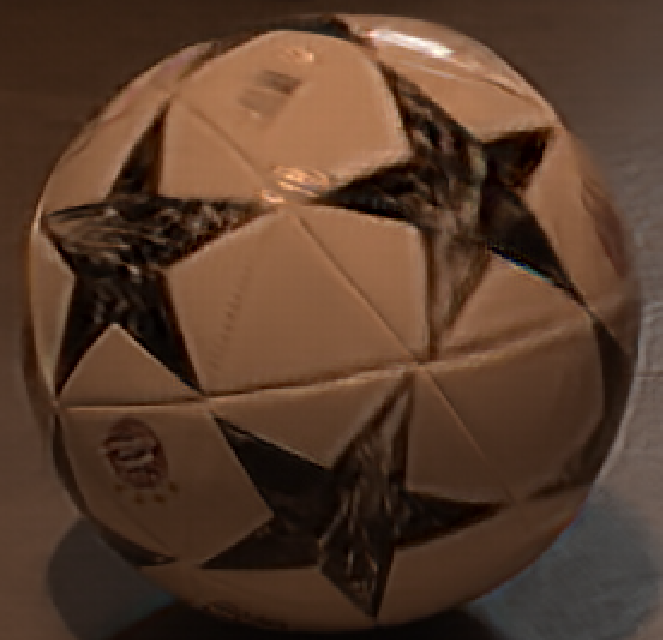}%
	\caption*{Frame 5}
	\end{subfigure}%
	\begin{subfigure}[b]{0.124\textwidth}
	\includegraphics[width=\linewidth,trim={0cm 0cm 7cm 0cm},clip]{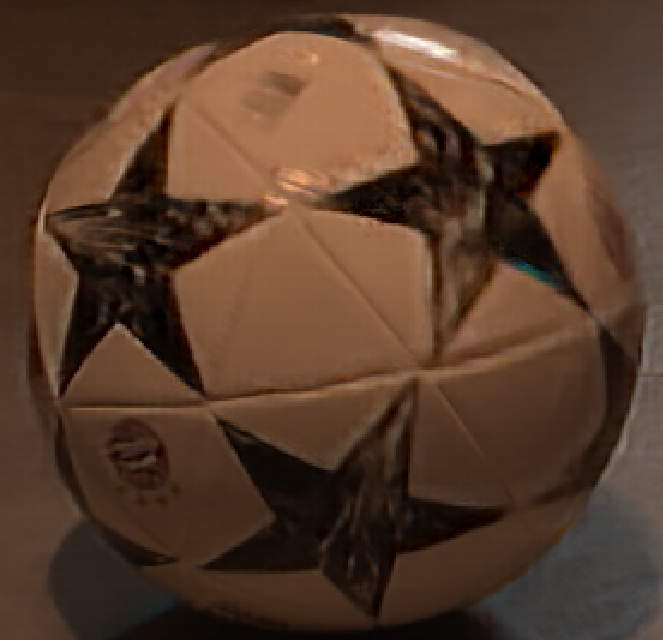}%
	\caption*{Frame 6}
	\end{subfigure}%
	\begin{subfigure}[b]{0.124\textwidth}
	\includegraphics[width=\linewidth,trim={0cm 0cm 7cm 0cm},clip]{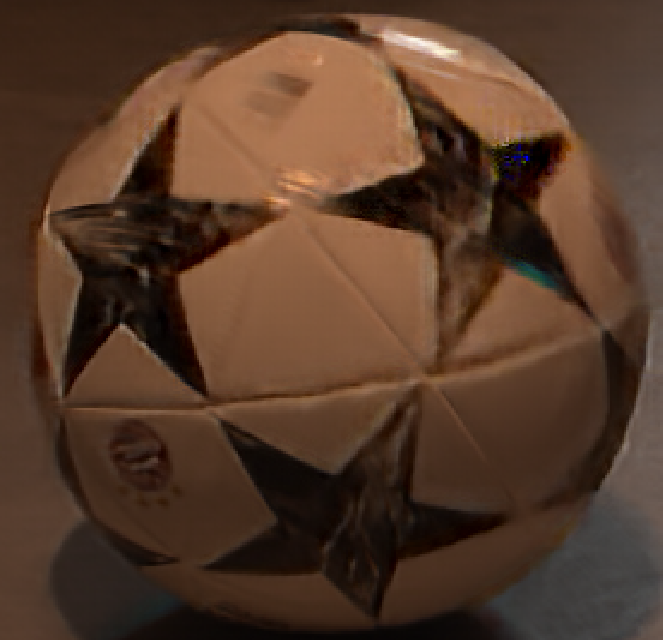}
	\caption*{Frame 7}
	\end{subfigure}	
	\caption{Examples with real images. All images are captured with a DSLR camera. 
	Top row: an image with multiple moving objects. Notice that the background scene is static. Second row Frame 1-7: the reconstructed video (requires zooming).
Rows 3 and 4: details of the scene.	Notice that the reconstruction shows both cars moving left to right. This is not the true motion (it would correspond to one vehicle reversing on the street). Bottom row: a rotating ball. The network can correctly reconstruct a video with a complex motion field. 
	}
\label{fig:realResults}
\vspace{-1.5em}
\end{figure*}

\noindent\textbf{Independent Frame Reconstruction. }
 A straightforward way of estimating a full sharp sequence is to replicate the training for each frame by minimizing the loss
$
{\cal L}_\text{indep}=\sum_{i=1}^{T} | \phi_i(y) - x[i] |^2,
\label{eq:trivialSolution}
$ 
where $\phi_i$ is a network to predict $x[i]$.
In this section we show that this scheme is not applicable beyond middle frames, due to the temporal ordering ambiguity. Fig.~\ref{fig:SchemeResults1} (a) and (g) show the visual results with an independent frame reconstruction scheme. From the results we can see that the quality of the reconstructed frame worsens as the distance from the middle frame increases. 

\noindent\textbf{Global Frame Reconstruction. }
In this section we show that the global ordering-invariant loss is not a good option either. Fig.~\ref{fig:SchemeResults1} (b) and (h) show the reconstructed 7 frames, where the middle frame is reconstructed independently with the loss $\cal L_\text{middle}$ and the other 6 frames are reconstructed jointly with the globally ordering-invariant loss $\cal L_\text{model}$. We can see that the non middle frame prediction network does not converge well and generates artifacts. 

\noindent\textbf{Pairwise Frame Reconstruction. }
Fig.~\ref{fig:SchemeResults1} (c), (d), (i) and (j) show the reconstructions with the pairwise ordering-invariant loss $\cal L_\text{pair}$. Rows (d) and (j) show the case where the middle frame prediction is also fed to the network, while the third row shows the case without the middle frame prediction. There are two main limitations of using $\cal L_\text{pair}$: 1) One has to manually reorder non middle frame predictions; 2) Although feeding the middle frame prediction to the network gives better visual results than in the case without it, still both of these two schemes generate artifacts, especially for the frames temporally away from the middle frame.

\noindent\textbf{Sequential Pairwise Frame Reconstruction. }
Fig.~\ref{fig:SchemeResults1} (e) and (k) show the visual results with a sequential pair-wise reconstruction scheme. Notice that this scheme and the pairwise ordering scheme only differ at the 4 frame predictions $x[1],x[2],x[6]$, and $x[7]$. We can see that the sequential scheme generates fewer artifacts especially at frames $x[1]$ and $x[7]$ in Fig.~\ref{fig:SchemeResults1} (k).

\noindent\textbf{Teacher Forcing. }
We also explore the \emph{teacher forcing} method used to train recurrent neural networks \cite{Goodfellow-et-al-2016}. During training we substitute the middle frame prediction $\phi_4(y)$ with the ground truth middle frame $x[4]$ in step 2 and 3 of our full training procedure in Table~\ref{tab:training}. This strategy brings several benefits: 1) In practice, we observe that the teacher forcing training strategy converges faster than with a standard sequential pairwise training; 2) It also gives visually better predictions as shown in Fig.~\ref{fig:SchemeResults1}~(l), where non middle frames are predicted by a network trained with teacher forcing; 3) The middle frame and non middle frame prediction training can be done in parallel. We use teacher forcing training as our default network training scheme.

We use 4 different networks to predict all $7$ frames: one to predict the middle frame, and the other three for middle-symmetric pair-wise frames. We found that sharing the parameters of the pair-wise networks for frames 1, 2, 6 and 7 during training would not result in a loss of visual accuracy. This could make predicting more than 7 frames feasible.

\noindent\textbf{Importance of the Middle Frame Estimate. }
We observe experimentally that a good initialization is key in making the non middle frame prediction network work well. 
Fig.~\ref{fig:FailResults} (a) shows a blurry image and an enlarged detail with significant motion blur. In Fig.~\ref{fig:FailResults} (b) we show the reconstructions of frame 1, 4 (middle) and 7 with our trained network, where we used our estimated frame 4 to recover the other frames. In Fig.~\ref{fig:FailResults} (c) we show the corresponding frames reconstructed when feeding our network with Nah's \cite{Nah_2017_CVPR} frame 4 estimate. Both cases fail to reconstruct the middle frames and the other frames. However, when we feed our networks with the ground truth middle frame (see Fig.~\ref{fig:FailResults} (d)), they are able to correctly reconstruct the other frames.

\noindent\textbf{Discussion. }
Although our system can predict 7 frames from a motion-blurred image, there are two main limitations.
One main limitation of our approach is that it is not robust to large blurs. Whenever our middle frame prediction network fails to remove blur, the non middle frame prediction networks also fail. 
\vspace{-0.5em}
\section{Conclusions} 
\vspace{-0.5em}
In this paper we have presented a first method to reconstruct a video from a single motion-blurred image. We have shown that the task is more ambiguous than deblurring a single frame because the temporal ordering is lost in the motion-blurred image. We have presented a data-driven solution that allows a convolutional neural network choose a temporal ordering at the output. We have demonstrated our model on several datasets and have shown that it generalizes on real images captured with different cameras from those used to collect the training set.

{\small\smallskip
\noindent\textbf{Acknowledgements.}
MJ, GM, and PF acknowledge support from the Swiss National Science Foundation on projects 200021\_153324 and 200021\_165845.
}

{\small
\bibliographystyle{ieee}
\bibliography{deblur}

\begin{thebibliography}{10}\itemsep=-1pt

\bibitem{Bahat_2017_ICCV}
Y.~Bahat, N.~Efrat, and M.~Irani.
\newblock Non-uniform blind deblurring by reblurring.
\newblock In {\em ICCV}, 2017.

\bibitem{Chakrabarti_2016_ECCV}
A.~Chakrabarti.
\newblock A neural approach to blind motion deblurring.
\newblock In {\em ECCV}, 2016.

\bibitem{ChoWL12}
S.~Cho, J.~Wang, and S.~Lee.
\newblock Video deblurring for hand-held cameras using patch-based synthesis.
\newblock {\em {ACM} Trans. Graph.}, 2012.

\bibitem{Dong_2017_ICCV}
J.~Dong, J.~Pan, Z.~Su, and M.-H. Yang.
\newblock Blind image deblurring with outlier handling.
\newblock In {\em ICCV}, 2017.

\bibitem{Fergus_2006_SIGGRAPH}
R.~Fergus, B.~Singh, A.~Hertzmann, S.~T. Roweis, and W.~T. Freeman.
\newblock Removing camera shake from a single photograph.
\newblock In {\em SIGGRAPH}, 2006.

\bibitem{Gong_2016_CVPR}
D.~Gong, M.~Tan, Y.~Zhang, A.~van~den Hengel, and Q.~Shi.
\newblock Blind image deconvolution by automatic gradient activation.
\newblock In {\em CVPR}, 2016.

\bibitem{Gong_2017_CVPR}
D.~Gong, J.~Yang, L.~Liu, Y.~Zhang, I.~Reid, C.~Shen, A.~van~den Hengel, and
  Q.~Shi.
\newblock From motion blur to motion flow: A deep learning solution for
  removing heterogeneous motion blur.
\newblock In {\em CVPR}, 2017.

\bibitem{Goodfellow-et-al-2016}
I.~Goodfellow, Y.~Bengio, and A.~Courville.
\newblock {\em Deep Learning}.
\newblock MIT Press, 2016.

\bibitem{Goodfellow_2014_NIPS}
I.~Goodfellow, J.~Pouget-Abadie, M.~Mirza, B.~Xu, D.~Warde-Farley, S.~Ozair,
  A.~Courville, and Y.~Bengio.
\newblock Generative adversarial nets.
\newblock In {\em NIPS}, 2014.

\bibitem{Hirsh}
S.~Harmeling, B.~Scholkopf, C.~J. Schuler, and M.~Hirsch.
\newblock Fast removal of non-uniform camera shake.
\newblock {\em ICCV}, 2011.

\bibitem{He_2016_CVPR}
K.~He, X.~Zhang, S.~Ren, and J.~Sun.
\newblock Deep residual learning for image recognition.
\newblock In {\em CVPR}, 2016.

\bibitem{He_2016_ECCV}
K.~He, X.~Zhang, S.~Ren, and J.~Sun.
\newblock Identity mappings in deep residual networks.
\newblock In {\em ECCV}, 2016.

\bibitem{Kim_2013_ICCV}
T.~Hyun~Kim, B.~Ahn, and K.~Mu~Lee.
\newblock Dynamic scene deblurring.
\newblock In {\em ICCV}, 2013.

\bibitem{Kim_2014_CVPR}
T.~Hyun~Kim and K.~Mu~Lee.
\newblock Segmentation-free dynamic scene deblurring.
\newblock In {\em CVPR}, 2014.

\bibitem{Kim_2017_ICCV}
T.~Hyun~Kim, K.~Mu~Lee, B.~Scholkopf, and M.~Hirsch.
\newblock Online video deblurring via dynamic temporal blending network.
\newblock In {\em ICCV}, 2017.

\bibitem{JohnsonAF16}
J.~Johnson, A.~Alahi, and L.~Fei{-}Fei.
\newblock Perceptual losses for real-time style transfer and super-resolution.
\newblock In {\em {ECCV}}, 2016.

\bibitem{Kim_2016_CVPR}
J.~Kim, J.~Kwon~Lee, and K.~Mu~Lee.
\newblock Deeply-recursive convolutional network for image super-resolution.
\newblock In {\em {CVPR}}, 2016.

\bibitem{Kim2016}
T.~H. Kim, S.~Nah, and K.~M. Lee.
\newblock Dynamic scene deblurring using a locally adaptive linear blur model.
\newblock {\em CoRR}, 2016.

\bibitem{Lai_2016_CVPR}
W.-S. Lai, J.-B. Huang, Z.~Hu, N.~Ahuja, and M.-H. Yang.
\newblock A comparative study for single image blind deblurring.
\newblock In {\em CVPR}, 2016.

\bibitem{Michaeli_2014_ECCV}
T.~Michaeli and M.~Irani.
\newblock Blind deblurring using internal patch recurrence.
\newblock In {\em ECCV}, 2014.

\bibitem{Nah_2017_CVPR}
S.~Nah, T.~Hyun~Kim, and K.~Mu~Lee.
\newblock Deep multi-scale convolutional neural network for dynamic scene
  deblurring.
\newblock In {\em CVPR}, 2017.

\bibitem{Nimisha_2017_ICCV}
T.~M. Nimisha, A.~Kumar~Singh, and A.~N. Rajagopalan.
\newblock Blur-invariant deep learning for blind-deblurring.
\newblock In {\em ICCV}, 2017.

\bibitem{debluringWild_2017_GCPR}
M.~Noroozi, P.~Chandramouli, and P.~Favaro.
\newblock Motion deblurring in the wild.
\newblock In {\em GCPR}, 2017.

\bibitem{Pan_2017_ICCV}
J.~Pan, J.~Dong, Y.-W. Tai, Z.~Su, and M.-H. Yang.
\newblock Learning discriminative data fitting functions for blind image
  deblurring.
\newblock In {\em ICCV}, 2017.

\bibitem{Pan_2016_CVPR_1}
J.~Pan, Z.~Hu, Z.~Su, H.-Y. Lee, and M.-H. Yang.
\newblock Soft-segmentation guided object motion deblurring.
\newblock In {\em CVPR}, 2016.

\bibitem{Pan_2016_CVPR}
J.~Pan, D.~Sun, H.~Pfister, and M.-H. Yang.
\newblock Blind image deblurring using dark channel prior.
\newblock In {\em CVPR}, 2016.

\bibitem{Pan_2017_CVPR}
L.~Pan, Y.~Dai, M.~Liu, and F.~Porikli.
\newblock Simultaneous stereo video deblurring and scene flow estimation.
\newblock In {\em CVPR}, 2017.

\bibitem{Park_2017_ICCV}
H.~Park and K.~Mu~Lee.
\newblock Joint estimation of camera pose, depth, deblurring, and
  super-resolution from a blurred image sequence.
\newblock In {\em ICCV}, 2017.

\bibitem{Ren_2017_ICCV}
W.~Ren, J.~Pan, X.~Cao, and M.-H. Yang.
\newblock Video deblurring via semantic segmentation and pixel-wise non-linear
  kernel.
\newblock In {\em ICCV}, 2017.

\bibitem{Sellent_2016_ECCV}
A.~Sellent, C.~Rother, and S.~Roth.
\newblock Stereo video deblurring.
\newblock In {\em ECCV}, 2016.

\bibitem{SimonyanZ14a}
K.~Simonyan and A.~Zisserman.
\newblock Very deep convolutional networks for large-scale image recognition.
\newblock In {\em {ICLR}}, 2015.

\bibitem{Su_2017_CVPR}
S.~Su, M.~Delbracio, J.~Wang, G.~Sapiro, W.~Heidrich, and O.~Wang.
\newblock Deep video deblurring for hand-held cameras.
\newblock In {\em CVPR}, 2017.

\bibitem{Sun_2015_CVPR}
J.~Sun, W.~Cao, Z.~Xu, and J.~Ponce.
\newblock Learning a convolutional neural network for non-uniform motion blur
  removal.
\newblock In {\em CVPR}, 2015.

\bibitem{Wieschollek_2017_ICCV}
P.~Wieschollek, M.~Hirsch, B.~Scholkopf, and H.~P.~A. Lensch.
\newblock Learning blind motion deblurring.
\newblock In {\em ICCV}, 2017.

\bibitem{Yan_2017_CVPR}
Y.~Yan, W.~Ren, Y.~Guo, R.~Wang, and X.~Cao.
\newblock Image deblurring via extreme channels prior.
\newblock In {\em CVPR}, 2017.

\bibitem{Wipf_2013_NIPS}
H.~Zhang and D.~P. Wipf.
\newblock Non-uniform camera shake removal using a spatially-adaptive sparse
  penalty.
\newblock In {\em NIPS}, 2013.

\bibitem{Zhang_2015_CVPR}
H.~Zhang and J.~Yang.
\newblock Intra-frame deblurring by leveraging inter-frame camera motion.
\newblock In {\em CVPR}, 2015.

\end{thebibliography}
}

\end{document}